\documentclass[11pt,letterpaper]{article}
\usepackage{cogsys}
\usepackage[T1]{fontenc}
\usepackage{times}
\usepackage[pdftex]{graphicx} %
\usepackage{rotating}

\usepackage{natbib}
\usepackage{algorithm}
\usepackage{algorithmic}
\usepackage{listings}

\ShortHeadings{Problem Specifications via Large Language Models}
              {R.E.\ Wray, J.\ Kirk, J.E.\ Laird}

\begin{document} 

\title{Eliciting Problem Specifications via Large Language Models}
 
\author{Robert E. Wray}{robert.wray@cic.iqmri.org}
\author{James R. Kirk}{james.kirk@cic.iqmri.org}
\author{John E. Laird}{john.laird@cic.iqmri.org}
\address{Center for Integrated Cognition, IQM Research Institute, 
         Ann Arbor, MI 48105 USA}
\vskip 0.2in

\begin{abstract}
Cognitive systems generally require a human to translate a problem definition into some specification that the cognitive system can use to attempt to solve the problem or perform the task. In this paper, we illustrate that large language models (LLMs) can be utilized to map a problem class, defined in natural language, into a semi-formal specification that can then be utilized by an existing reasoning and learning system to solve instances from the problem class. We present the design of LLM-enabled \textit{cognitive task analyst} agent(s). Implemented with LLM agents, this system produces a definition of problem spaces for tasks specified in natural language. LLM prompts are derived from the definition of problem spaces in the AI literature and general problem-solving strategies (Polya's \textit{How to Solve It}). A cognitive system can then use the problem-space specification, applying domain-general problem solving strategies (``weak methods'' such as search), to solve multiple instances of problems from the problem class. This result, while preliminary, suggests the potential for speeding cognitive systems research via disintermediation of problem formulation while also retaining core capabilities of cognitive systems, such as robust inference and online learning.
\end{abstract}

\section{Introduction}

From the earliest days of Artificial Intelligence as a discipline of study, researchers have sought to enable AI systems -- agents -- to attack general problems where no known/existing special-purpose method is known. Examples of these methods include various kinds of systematic search \citep[e.g., hill climbing, depth- and breadth-first search,][]{newell_computer_1976,nils_nilsson_problem_1971}, 
planning \citep{fikes_strips_1971}, means-ends analysis \citep{newell_gps_1963}, and general heuristics intended to reduce the total size of search spaces, including some inspired by human problem solving \citep{newell_human_1972}. These \textit{weak methods} \citep{newell_heuristic_1969,laird_universal_1983}
remain critical in state-of-art agents today. For example, both Q-learning \citep{sutton_reinforcement_1999} 
and backpropagation \citep{rumelhart_learning_1986} 
use a form of hill-climbing search..

A limitation of weak methods, recognized even in these early days of AI, is the human interpretation required to translate from a problem description to a formal specification that can be used by a weak method. In discussing the limitations of weak methods as a general foundation for solving ``ill-structured problems'' (essentially meaning classes of problems where no specialized method is known to exist), Newell \citeyearpar{newell_heuristic_1969} relates: 
\begin{quote}
    [R]epresentation of problems also raises a question of the locus of power.... we talk about the representation of a problem ... presumably a translation from its representation in some other form, such as natural language. These changes of the basic representational system are clearly of great importance to problem solving.... A suspicion arises that changes of the representation at this level ... might constitute a substantial part of problem solving.
\end{quote}

Then and now, a human generally is required to perform this ``translation'' (e.g., knowledge engineering, reward design, feature selection) from the problem to its representation in an AI system. In the intervening years, methods that systematized and codified these translation processes have arisen in knowledge-based systems \citep{schreiber_knowledge_2000,arp_building_2015},
 engineering psychology  \citep{crandall_working_2006,schraagen_cognitive_2000},
and machine learning \citep{langley_selection_1994,cai_feature_2018},
illustrating that, indeed, the initial representation of the problem is a significant locus of power toward solution.

Today, large language models (LLMs) are being evaluated for and (sometimes) used for analytic tasks including software development \citep{hou_large_2024,ozkaya_application_2023},
general research review and synthesis \citep{shao_assisting_2024}, 
and literature/text analysis \citep{tornberg_how_2023,gilardi_chatgpt_2023}. 
The most recent versions of LLMs allow users to create virtual assistants customized to specific classes of tasks \citep{dong_towards_2023}, enable the use of external tools for specialized tasks \citep{kim_language_2023,xu_rewoo_2023,qin_tool_2023}, 
and support contextual, dynamic, agent-like decision making \citep[``agentic'' LLM systems, e.g.,][]{yao_react_2023,wang_survey_2024}.

Given these LLM capabilities, is it feasible for an LLM, when presented with a natural language problem description, to automatically produce  ``translations'' -- formal specifications of the problem space and problems -- that enable the immediate application of weak methods? Below, we propose and define a virtual assistant with the role of \textit{cognitive task analyst}. The virtual assistant uses a collection of domain-general prompts to elicit information about the problem. More specifically, prompts are designed around the concept of problem spaces \citep{newell_reasoning_1980} as the target of a problem specification. The prompts also draw from \textit{How to Solve It} \citep{polya_how_2015}, a well-known/classic treatise on general methods to attack unfamiliar problems.

\begin{table}[bth]
    \centering
    \caption{Natural language descriptions of problem space and problem from Newell \citeyearpar{newell_reasoning_1980}. These descriptions are provided \textit{verbatim} to the LLM in the prompts and prompt templates described below.}
    \begin{small}

    \begin{tabular}{p{.10\textwidth}p{.85\textwidth}}
    \hline
         \textbf{Problem Space}: &  A problem space consists a set of symbolic structures (the states of the space) and a set of operators over the space. Each operator takes a state as input and produces a state as output (although there may be other inputs and outputs as well). The operators may be partial (i.e., not defined for all states). Sequences of operators define paths that thread their way through sequences of states. \\
         \textbf{Problem}: & A problem in a problem space consists of a set of initial states, a set of goal states, and a set of path constraints. The problem is to find a path through the space that starts at the initial state, passes only along paths that satisfy the path constraints, and ends at any goal state. \\ \hline
    \end{tabular}
     \end{small}
    \label{tab:probspace-defn}
\end{table}

\section{Background: Problem Spaces and Cognitive Task Analysis}

In this section, we briefly summarize what we mean by ``problem space'' and enumerate various steps and outputs that are needed to formulate a problem. This analysis provides requirements for the problem-space formulation (``translation'') processes outlined above.

Table~\ref{tab:probspace-defn} presents Newell's \citeyearpar{newell_reasoning_1980} definitions of problem space and problem. A problem space (generally) defines how a number of different problems of the same class can be approached, outlining what situations can be encountered (states) and what actions are available in those situations (operators). 
For example, a problem space for basic integer arithmetic might include single-digit addition and subtraction operators, a carry operator, and integers (states). The problem space can then be used to express (and solve) individual problem instances (i.e., $113-67$). As another example, consider the classic ``water jugs'' problem in which a specific amount of water must be produced in one container from two  containers that do not support any partial measurement (and an unlimited supply of water). The problem-space operators and states for this problem are defined in Table~\ref{tab:probspace-example}.

\begin{table}[h]
    \caption{Definition the Problem Space for Water Jugs from \citep{laird_universal_1983}.}
    \centering
    \begin{small}
    \begin{tabular}{p{.2\textwidth}p{.75\textwidth}}
    \hline
         \textbf{State}: &  Current volume of each jug. 
         \\
         \hline
         \textbf{Operators}: & \textbf{Fill} a given jug. ~~~~  \textbf{Empty} a given jug. \\
         & \textbf{Pour} the contents of jug into the other until the source jug is empty or the receiving jug is full.\\
         \hline
         \textbf{Path Constraints}: & If the target jug has the goal amount, done.\\
         & Do not undo the previous action. \\
         & Do not take an action that will produce a state already on the path (avoid loops) \\
         \hline
    \end{tabular}
    \end{small}

    \label{tab:probspace-example}
\end{table}

Path constraints define allowed orders of operations. For example, in basic arithmetic, there is (typically) a path constraint that the one's column be computed prior to the ten's column. 
When a specialized procedure is not known to the solver, search can be used to find a solution to a problem within the problem space. The path constraints for ``water jugs'' (Table~\ref{tab:probspace-example}) are much less specific than the rigid constraints for arithmetic procedures. However, they do constrain the resulting search, avoiding inverse actions and loops that would occur in search otherwise.\footnote{Path constraints and search control (discussed later) both influence the sequence of operations. Generally, path constraints prescribe what actions are legal and not and search control guides what choices are more/less preferable.} 

Problem spaces describe the components of problem solving. Where do such specifications come from?
As suggested above, various methods of task analysis have been developed that allow analysts to document how problems should be solved. Cognitive task analysis \citep[CTA,][]{crandall_working_2006} is one method that guides an analyst in interviewing a human expert to capture and to codify how they approach or perform various tasks. We choose CTA because the design of an early and influential form of CTA, GOMS \citep{john_goms_1996}, focused analysis somewhat directly on the formulation of problem spaces compatible with Newell's definition. GOMS guides the analyst in identifying Goals, Operators, Methods, and Selection knowledge (roughly, path constraints) for a task. Below, we take advantage of this connection by defining a ``Cognitive task analyst'' role for the LLM agent. By creating this role context for the LLM, the intent is to bias generation toward concepts consistent with this conception of problem spaces (and their requirements).

CTA is generally focused on eliciting knowledge where a solution method exists/is known. For our research, we do not want to assume that a method is ``known'' to the LLM. The LLM can still take advantage of any relevant and known methods, but more general strategies are likely needed as well. For such ``general problem-solving strategy'' knowledge, we include in the design of the CTA Agent prompts derived from \textit{How to Solve It} \citep{polya_how_2015}. This book outlines an abstract, step-wise process for solving unfamiliar problems, including defining knowns, unknowns, goals, etc. One of the important ideas from Polya that we embed in the CTA Agent is the idea of introducing notation to define states and operators. While LLMs are generally viewed as generators of natural language, research has shown that they can also be used to produce specifications in formal/semi-formal notations \citep{ye_generating_2023,ye_satlm_2023}. It should also be noted that while LLMs can generate such specifications, they are somewhat less able to (natively) manipulate statements in a formal notation. %

\section{Overall System Design}

\begin{figure}[t]
\vskip 0.05in
\begin{center}
\includegraphics[width=.75\textwidth]{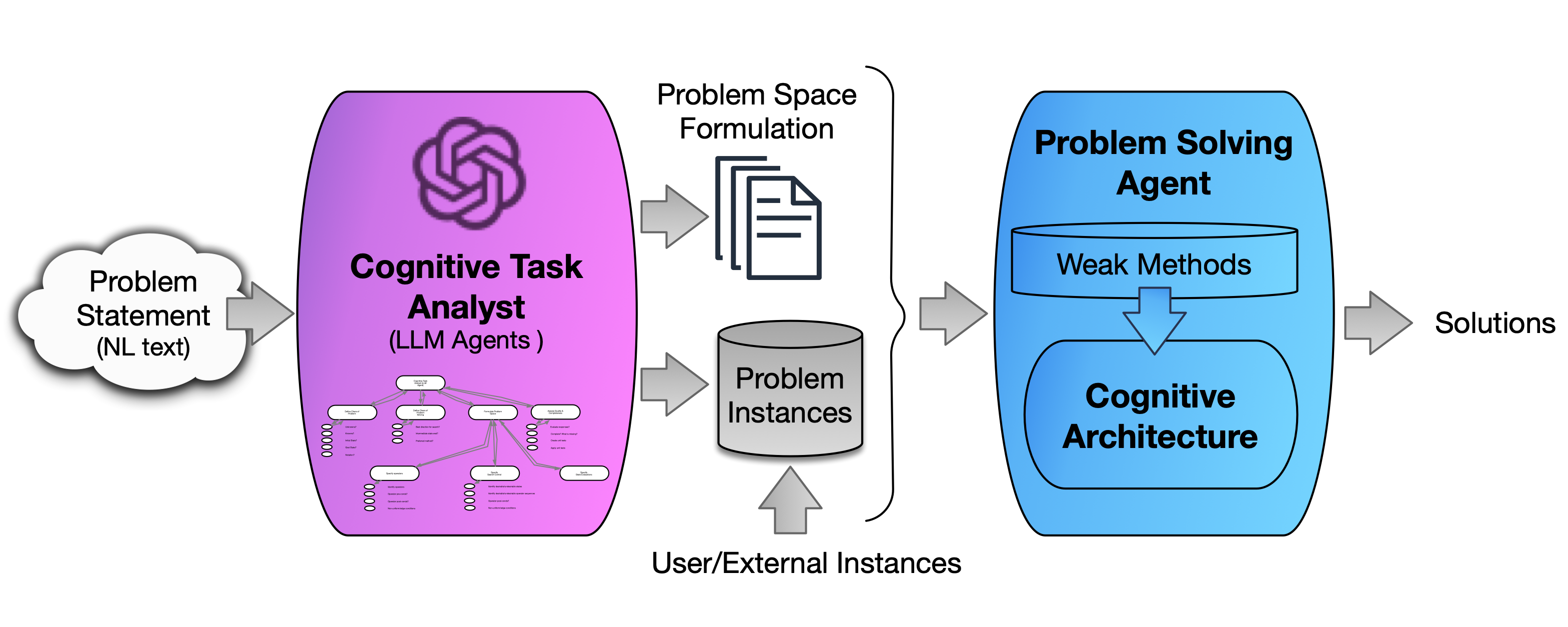}
\caption{High-level systems architecture comprising the ``CTA'' agent and a cognitive architecture.} 
\label{fig:systems-diagram}
\end{center}
\vskip -0.2in
\end{figure}

Figure~\ref{fig:systems-diagram} presents an overall systems design for the envisioned problem-solving system. A problem description is presented in natural language to the CTA Agent.\footnote{For this feasibility exploration, we focus solely on text. Future elaborations of this architecture might include multimodal inputs, given the emergence of multimodal LLMs.} Task analysis is conducted by the ``Cognitive Task Analyst'' agent, which is implemented as a collection of LLM agents. These agents, discussed further in the next section, are designed to conduct various parts of analysis, such as specifying operators or performing quality assurance checks on intermediate results. 

The output of the CTA Agent is a specification of the problem space and a collection of problem instances. Minimally, the specification of the problem space will include a declaration of the features relevant to problem solving (state description, path constraints), declaration of the applicable operators, and search control.  Determining the most appropriate ``form'' of the specification is a subject of future work. Long-term, the form might be some kind of declarative language as in SATLM \citep{ye_satlm_2023} or PDDL \citep{fox_pddl21_2003} or we could investigate direct code generation. Because we are focused in this paper on feasibility evaluation, we require for now that the form can be expressed in a consistent notation that can be directly mapped (by a human) to knowledge components of the cognitive architecture.

Problem instances can be defined by both the CTA Agent and human users. For example, in cognitive task analysis, it is common for the analyst to generate common and ``edge'' cases, the purpose of which is to test the correctness and sufficiency of the (in progress) problem-space formulation. The CTA Agent should thus also have such capability. However, we also want to test problem solving on problem instances that the LLM did not explicitly consider in formulating the problem space. Thus, the design anticipates that human users may also add problem instances.

Finally, the problem-space formulation is used by the problem-solving agent to produce solutions for individual instances of the problem. For this paper, we assume that the problem-solving agent consists of only the fixed mechanisms of the agent architecture and a store of domain-general weak methods. The weak methods enable search for a solution within the defined problem space. 

We hypothesize that the problem-solving agent will be able to use the problem-space formulation and the weak methods to solve (many) problem instances without needing human intervention or input for either problem-space formulation or additional knowledge (beyond the weak methods) in the problem-solving agent. A secondary question is to assess the extent to which the cost of solutions is similar to human-formulated problem spaces. For instance, the CTA Agent's problem-space formulation might correctly enumerate operators and states but omit search control, resulting in a much larger search space than the human-created problem-space specification. In other words, agent-generated problem-space formulations that require many orders of magnitude more steps to solve than human-formulated ones would only partially satisfy the main hypothesis.

\begin{figure}[tb]
\vskip 0.05in
\begin{center}
\includegraphics[width=.75\textwidth]{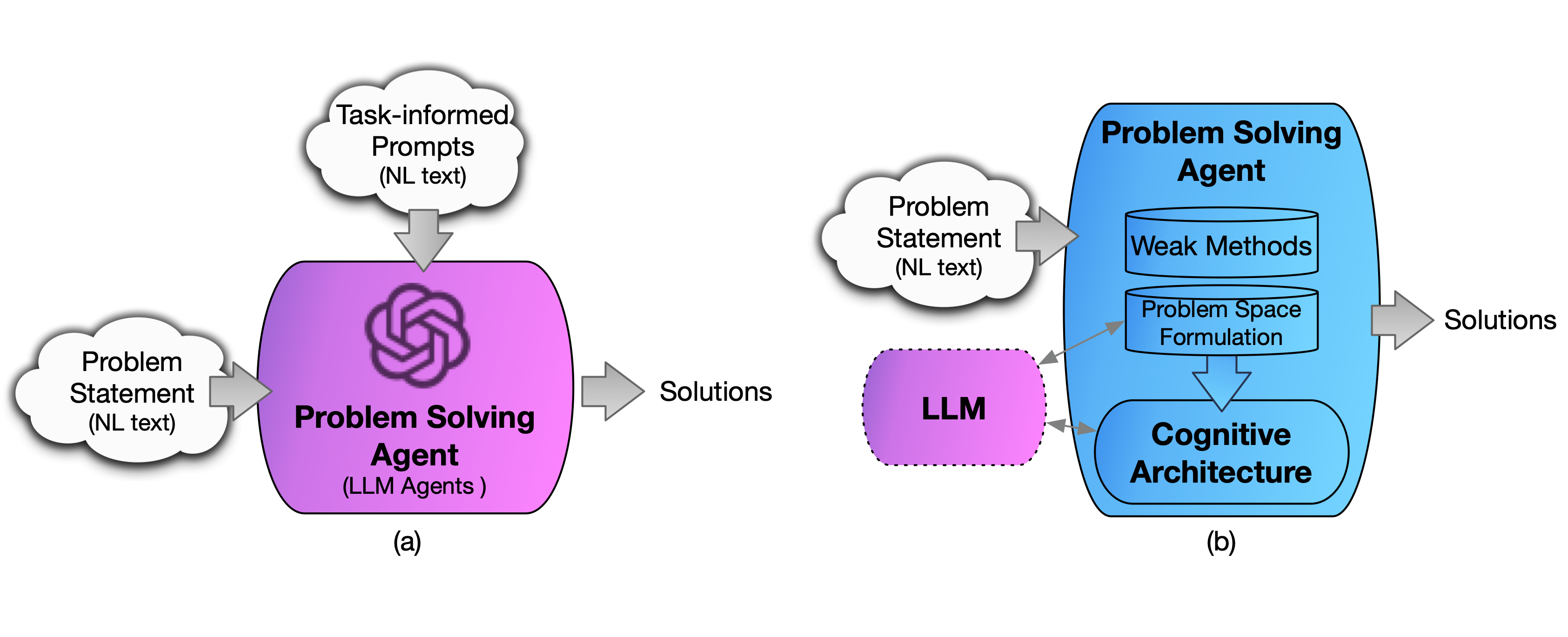}
\caption{Design space alternatives for a problem-formulation and problem-solving agent.} 
\label{fig:alternative-systems}
\end{center}
\vskip -0.2in
\end{figure} 

Before continuing with the design of the CTA Agent and exploration of these hypotheses, we briefly discuss two alternative systems designs, illustrated in Figure~\ref{fig:alternative-systems}. One option, illustrated in \ref{fig:alternative-systems}(a), would be to use primarily LLM components to solve the problem. We include in this alternative not only prompt-engineering strategies such as chain-of-thought \citep{wei_chain--thought_2022} and self-consistency, \citep{wang_self-consistency_2023}, 
but also agentic LLM approaches such as Tree of Thoughts \citep[ToT,][]{yao_tree_2023} and Graph of Thoughts \citep{besta_graph_2024}. For example, Tree of Thoughts attempts to solve problems similar to those we consider herein by having the language model itself generate search states. State descriptions, operators, goal evaluations, and search control are all implemented via prompts within the language model. ToT prompts and responses generally use natural language generation and interpretation (rather than formal representations or notation). ToT employs a simple memory, a stack of contexts, to enable backtracking in the search. However, Tree of Thoughts uses manually created, task-informed prompts to support problem-space search, in essence requiring human mediation for problem-space formulation. In contrast, our approach seeks to enable general problem-solving across many tasks without any task-specific prompting (or other human mediation), pre-training, or fine-tuning for the domain.

Another approach, as in \ref{fig:alternative-systems}(b), would be to create or to learn problem-space formulation knowledge within the agent itself. This approach is similar in spirit to work in Soar on Interactive Task Learning \citep[ITL,][]{kirk_learning_2016,mininger_expanding_2021}, where an agent learns to perform new tasks (playing games, various robotic tasks) via a series of interactions with a human to learn goals, constraints, and actions. In recent work, this ITL approach has been extended to integrate with an LLM \citep{kirk_improving_2024}. Rather than relying solely on human instruction, the agent extracts task knowledge from the LLM (e.g., learning what to do with various items like an empty can or dirty plate when tidying a kitchen). Integration with the LLM lessens what human feedback is needed for learning everyday, familiar tasks while maintaining solution quality.

Two related but distinct requirements distinguish our
goals from those of ITL broadly. First, we seek to remove human mediation from problem solving, going directly from problem description to solution. ITL, by design, depends on human input. Second, we seek to enable a problem solver to attack problems where no human already understands or can formulate an apt problem space for search (including the absence of solution(s) in the training data of the language model). In the long term, we see ITL and this work as complementary. When an agent has access to a human or existing domain-specific resources, it can draw on those to solve a problem, using capabilities such as those developed for ITL. When humans are not available and/or little/no prior information about the problem is known, the agent can still attempt to make progress by formulating a problem space and using its weak methods to attempt to search for potential solutions.

Finally, we observe that the application of LLMs to problem specification has been previously explored, most notably by \cite{valmeekam_planning_2023}. They compared the results of native LLM problem-solving against the results obtained from problem specification (where the LLM generated PDDL as output) and the subsequent application of a planning system using those specifications in the blocks world domain. Consistent with our hypothesis, the combination of LLM-mediated problem specification and the planner significantly outperformed the LLM solution alone. Additionally, they demonstrated that plan-repair systems could consistently improve LLM plan specifications that were incomplete or incorrect. 

A key distinction in our ambition and their approach is that their approach provided a (natural language) description of the problem space. For instance, for the blocks world, a user-authored prompt describes the operators in the domain and their effects. Our goal is for an LLM agent to generate these descriptions of operators and their effects itself. However, their overall results do provide compelling evidence of potential benefits of combining LLM-generated problem specifications and planning/search tools.

\section{Design of the Cognitive Task Analyst}

\begin{figure}[t]
\vskip 0.05in
\begin{center}
\includegraphics[width=.85\textwidth]{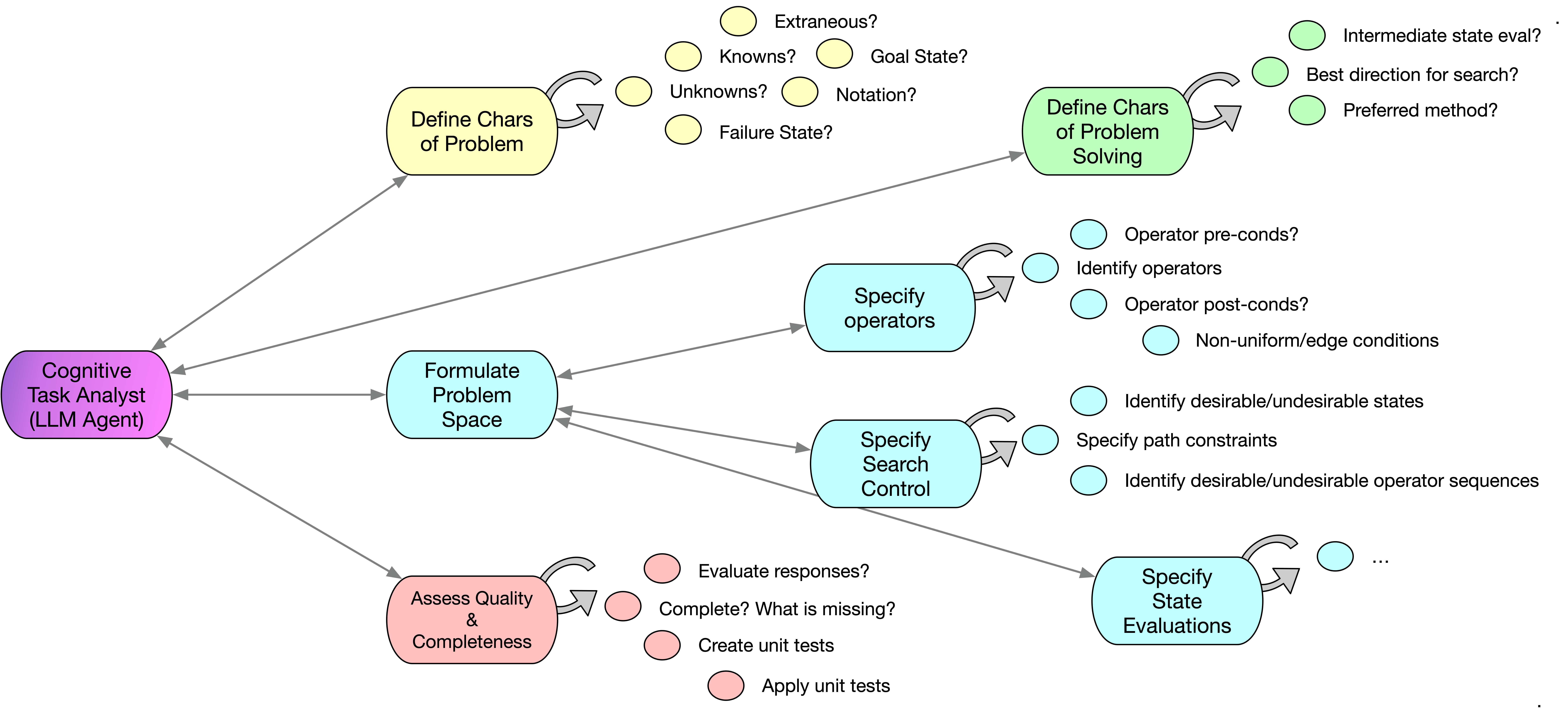}
\caption{Initial design of the Cognitive Task Analyst (CTA) LLM agent.} 
\label{fig:cta-agent}
\end{center}
\vskip -0.2in
\end{figure}

An overall design of the CTA Agent is illustrated in Figure~\ref{fig:cta-agent}. As task analysis proceeds, the  
agent chooses among four high-level directions in which to continue its analysis: 1) defining the characteristics of the problem, 2) defining characteristics of problem solving, 3) formulating the problem space, and 4) assessing the quality of the solution thus far. Within each of these four directions, more specific analysis questions (represented as nodes and subnodes in the figure) further decompose problem-space formulation. 

Leaf nodes are implemented as prompts for specific contributions to the task analysis. For example, a prompt might ask the LLM to identify the unknowns in the problem-space formulation or specify the preconditions of an operator (see next section for detailed examples). The result from each prompt/response is added to a memory that represents the problem-space formulation thus far. The contents of this memory are presented to the LLM with each individual prompt, creating an evolving ``context'' for each query. 

The arrows in the diagram are intended to convey the agent's ability to direct its analysis based on results thus far. Long-term, we envision an agent that can traverse this tree in many different ways, guided by the results of analysis thus far. For one problem definition, it might explore in a depth-first fashion (exploring problem characteristics first and then moving to problem-solving characteristics), in another a more breadth-first fashion (e.g., switching between problem characterization and problem-solving characterization), etc. In the implementation we describe below, the path(s) through the tree is fixed in order to best support feasibility analysis (e.g., comparison of results; repeatability).

A decision graph such as the one illustrated in the figure can be implemented using existing LLM development tools, such as LangGraph\footnote{https://python.langchain.com/docs/langgraph} or CrewAI\footnote{https://crewai.io/}. These tools support conditional execution of nodes, variable passing between nodes \citep{xu_rewoo_2023}, and dynamic reflection \citep{shinn_reflexion_2023}. Reflection, or assessing intermediate results, is particularly important. It is represented in the figure by the quality assurance direction of analysis, reflecting on what has been produced thus far. 

In the next section, we illustrate what is input to and produced by a few of these nodes with specific examples of task analysis based on variations of the water jug task. However, before turning to implementation results, we identify a few open/unresolved questions for a future, refined implementation that motivate our goals of this initial feasibility implementation.

    \begin{itemize}
        \item Generation for novel problems. An assessment of overall feasibility of the approach must attempt to evaluate the ability of the CTA Agent to formulate problem spaces for problems not in its training set. Variations include problems that are wholly novel and also problems that share some features with familiar problems but that are novel in whole. In particular, this last class of problem requires that the LLM generate good, novel analysis while mitigating the general tendency of LLMs to reproduce or ``parrot'' similar problems and solutions that appeared in its training set. As we show further below, we introduce some variations of the familiar water-jug task to begin to assess this question.
        \item Individual vs. distinct personas. While we envision the CTA Agent as having an analyst role or persona, developing distinct personas for LLM agents can be beneficial for improving  performance. For example, STORM, a system that generates text comparable to Wikipedia pages, implements separate editor and expert agents in the process of article development \citep{shao_assisting_2024}. Similarly, we might implement the quality assessment part of the CTA Agent to take on a quality-assurance (QA) engineer role (rather than the analyst role). Embedding reflective/QA functions in the feasibility implementation will help us identify gaps or unmet needs.
        \item Integrated vs. distinct analytic strategies. Is the analytic process sufficiently similar across different classes of problems that a single analytic strategy (as illustrated in        
        Figure~\ref{fig:cta-agent}) is sufficient?  This approach assumes that the task of defining a problem space for a DFS/BFS problem (like water jugs) significantly overlaps with other problem-space classes. For instance, defining a hill-climbing search problem space requires identifying a function to measure partial progress. A means-end approach requires identifying differences, neither of which is required for DFS/BFS problems.
        \item Means of information transfer. In the long term, a complete implementation of the CTA Agent should produce ``execution ready'' problem-space formulations. However, there are many potential targets for expressing the problem-space formulation, from direct generation of code (e.g., expressions of the problem space in Soar language) to stand-alone formal specification languages %
        \citep[e.g., PDDL,][]{green_what_2024,fox_pddl21_2003}. An LLM can successfully generate PDDL from natural language \citep{miglani_nltopddl_2020,valmeekam_planning_2023}. However, from our experiences creating translators from PDDL to Soar, we know that formal languages can impose representational and processing assumptions that do not take full advantage of architectural capabilities. For the initial implementation, we focus on having the CTA Agent generate a formal notation for the problem-space formulation and then manually implement that formal notation directly in Soar in order to better understand and evaluate the trade offs in direct encoding versus encoding to an intermediate form such as PDDL. 
    \end{itemize}

\begin{figure}[t]
\vskip 0.05in
\begin{center}
\includegraphics[width=.75\textwidth]{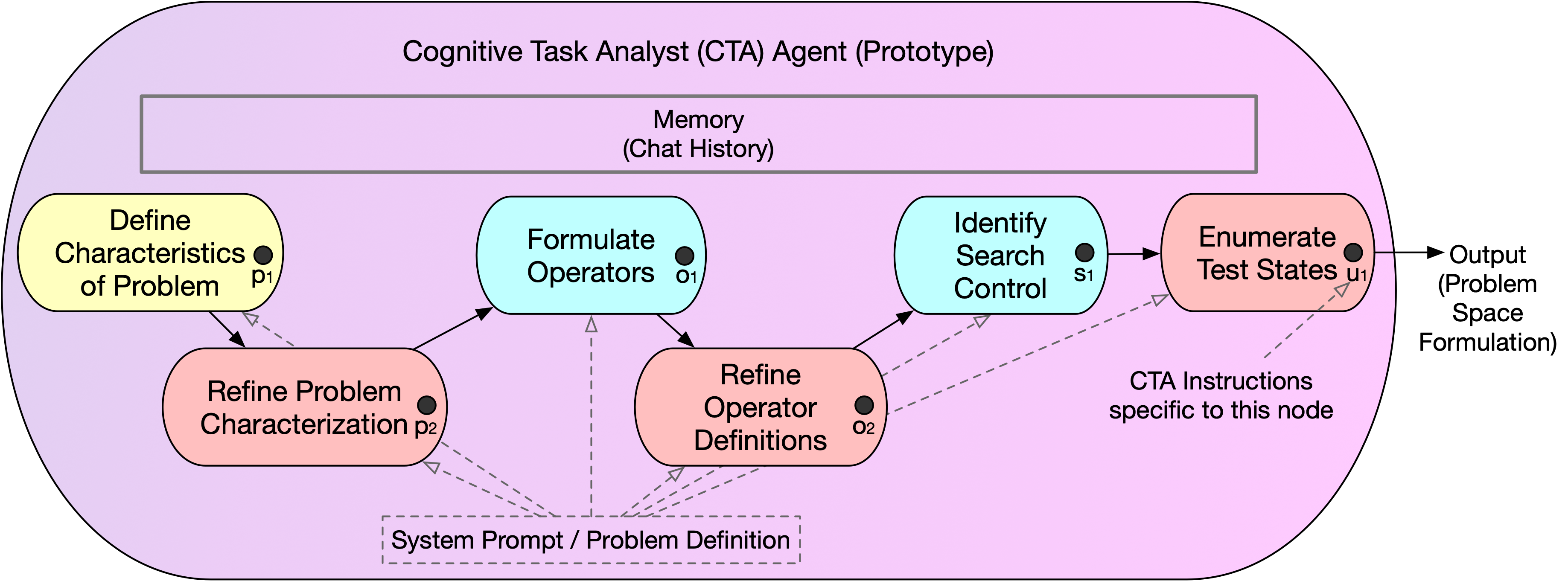}
\caption{Feasibility implementation of the Cognitive Task Analyst (CTA) LLM agent.} 
\label{fig:cta-agent-feasibility-implementation}
\end{center}
\vskip -0.2in
\end{figure}

\section{Feasibility Implementation and Evaluation Plan}

We implemented an initial version of the CTA Agent with LangGraph.\footnote{In a technical sense, the CTA Agent is implemented as a collection of LLM Agents in LangGraph. That is, the ``CTA Agent'' is implemented as a number of different, interoperating LLM agents, sometimes called a ``crew.'' We refer to the entire application as an ``agent'' in this paper and the individual LLM~agents as ``nodes.''} As mentioned above, in this initial feasibility implementation, the actual execution path through the network is sequential, following a fixed path we defined. Additionally, because we are limiting our exploration in this paper to one problem domain (water jugs) and variations of it, we do not introduce the questions relevant to hill-climbing domains because water jugs does not have an (obvious) measure of progress. Similarly, we omit problem-solving characterization because all the examples we explore are solvable with basic search.

Figure~\ref{fig:cta-agent-feasibility-implementation} summarizes the implemented agent. The agent first seeks to characterize the problem, then reflects/refines the characterization, then moves to problem-space formulation, search control, and finally to the identification of potential states/situations to test the system. The overall (system) prompt and the initial and refinement prompts for problem characterization are shown in Table~\ref{tab:feasiblity_prompt_examples}. Each node is implemented using a consistent pattern: the general context (system) prompt, a problem description, and specific instructions for that agent (e.g., instructions to ``refine the prior formulation,'' represented as filled circles in each node). All node prompts are specific to cognitive task analysis but do not mention or reference the problem. Variabilization of the problem description in each node allows the agent to generate alternative problem-space formulations for variations in the problem description, as we have detailed below.

\begin{table}[tb]
    \caption{Examples of prompts used in the CTA~Agent feasibility prototype.}
    \centering
    \begin{small}

    \begin{tabular}{p{1in}p{4.65in}}
    \hline
    Prompt Type & Prompt \\
    \hline
         System Prompt &  You are an expert in cognitive task analysis. 
You are helping to design a reasoner/problem solver that can solve many different instances of a class of problem. 
Problems are defined in 1-2 paragraphs. There will often be a specific example problem provided. 
However,  your responses should focus on contributions to the general formulation of the problem space. 
Focus each response on the most recent, specific question asked of you. 
The questions are designed to break the problem-space formulation into a set of discrete steps. 
Unless otherwise directed, be concise in each response (i.e., respond in one sentence or expression).\\ \hline
         Problem \par Characterization & For this response, focus on characterizing the problem itself. 
Use notation if/as possible to characterize the problem. Be concise.
What is the initial state? 
What is the final state? 
Are there illegal / impossible states that are not allowed for this problem? If so, identify them.\\ \hline
         Refinement of \par Problem \par Characterization & You specialize in reviewing the work of other analysts.
Given the PROBLEM DESCRIPTION and the previous AI response characterizing the problem, your task is to improve the prior problem characterization.
Are elements incorrect? If so, correct them.
Are elements missing? If so, add them. Focus on problem characterization only (specifications of states) rather than what actions to take.
Are elements poorly formed or ambiguous? For example, replace qualitative terms such as less or more with specific expressions that reflect quantitative values. 
Respond with a revised problem characterization that reflects your analysis. 
Use notation to characterize the problem. \\ \hline
    \end{tabular}
      \end{small}
    \label{tab:feasiblity_prompt_examples}
\end{table}

In this initial version of the agent, all previously generated responses are provided to each subsequent agent. That is, each subsequent node sees all the previous responses as context for its response. In future work, we will implement a special-purpose memory to hold the overall problem characterization and problem-space formulation; we did not find that necessary for this exploration. 

Obviously, water jugs is a familiar problem, and the LLMs we are using have examples of water jugs in their training set. However, because the training sets of LLMs are vast, it is generally difficult to ensure that any known problem is not already represented. Further, known problems might bias the LLM's response to responding appropriately to slight variations in a problem statement (``strong attractor'' effect). To explore these issues, we developed a number of distinct descriptions of water-jug problems for this feasibility exploration. We introduce the following notation to summarize specific water jug problems: $F|V|A(V_{s},...,Vol_{l})\rightarrow Vol_{goal})$. The initial letter represents the case type (Familiar, Variation, or Analogue). Table~\ref{tab:feasiblity_problem_descriptions} shows one example from each of the three case types of a specific problem description prompt provided to the LLM.
\begin{enumerate}
\item Familiar/Classic: These examples are intended to reflect examples likely in the training set data. We use the exact wording of the water jugs problem as it appears in \textit{How to Solve It} ($F(4,9)\rightarrow6$), a specific instance of water jugs that appeared in the movie \textit{Die Hard with a Vengeance} ($F(3,5)\rightarrow 4$), and an instance of the familiar problem with possibly novel (or at least uncommon) jug amounts ($F(9,17)\rightarrow 5$).
\item Variations of water jugs: These examples are intended to  trigger known solutions that are inapt for this problem space. We consider two specific variations:
\begin{itemize}
    \item Different units: The wording is the same as in the familiar examples, but the units differ between jugs ($V(4 qt, 9 gal)\rightarrow 6 gal$)
    \item Additional jugs: This problem introduces three rather than two jugs ($V(2,3,5)\rightarrow4$). 
\end{itemize}
\item Analogue: This case is meant to share structural identity with the familiar cases but little surface similarity; e.g., ($A(4,9)\rightarrow 6$). We use ``analogue'' to describe these, rather than ``disguised'' \citep[e.g., as in ][]{valmeekam_planning_2023}, because there is an overall consistent semantics for the analogue case, rather than word substitution.
\end{enumerate}

\begin{table}[t]
\caption{Alternative problem descriptions used to explore and to assess feasibility of the CTA~Agent.}
    \centering
    \begin{small}

    \begin{tabular}{p{\linewidth}}
    \hline
         \textbf{Familiar}:  The general problem is to deliver a specific amount of water using containers that are opaque and have contain no graduated markings. 
The amount of water that needs to be delivered will generally differ from the full capacities of the container. As a specific example, 
how can you bring up from the river exactly 6 quarts of water when you have only two containers, 
a four quart pail and a nine quart pail to measure with? \\ \hline
         \textbf{Distractor Variation (Different Units)}: The general problem ... As a specific example, 
how can you bring up from the river exactly 6 gallons of water when you have only two containers, 
a 4-quart pail and a 9-gallon pail to measure with?\\ \hline
         \textbf{Same Structure, Novel Surface Features}:  You are an engineer on an alien spaceship. 
The general problem is to deliver a specific amount of ``flucotone'' (a type of energy) using flucotone-holding devices (FHDs). 
Flucotone can be transferred to FHDs up to their specified capacity with no energy/flucotone loss. 
An FHD can hold any amount of flucotone up its capacity but there is no way to measure how much flucontone is in an FHD other than it being full or empty. 
The amount of flucotone needed for any specific task will generally differ from the full capacities of available FHDs. As a specific example, 
assume a power plant that can generate limitless flucotone. From this power plant, how can you deliver exactly 6 units of flucotone to the engine room  
when you have only two FHDs: a 4-unit FHD and a 9-unit FHD?\\ \hline
    \end{tabular}
    \end{small}
    \label{tab:feasiblity_problem_descriptions}
\end{table}

\section{Feasibility Evaluation}

We ran all the variations above with the CTA Agent using GPT3.5 (gpt-3.5-turbo-0125) and GPT4 (gpt-4-0125-preview) with temperature=0.\footnote{Temperature parameterizes the variability of sequential token generation by the LLM. Temperature=0 minimizes variability. However, due to general indeterminism in token generation, temperature=0 responses will sometimes differ in response to the same prompt and context. For this feasibility exploration, we generally ran the LLM once for each condition and thus did not explore potential variability. The total cost of the API tokens (for both models) for the work reported in this paper was $\sim$\$3USD.} For comparison, we also ran a one-shot problem-space formulation using a chain-of-thought instruction \citep{wei_chain--thought_2022}. We also presented LLM with a request to solve each problem instance directly (also using a chain-of-thought prompt). In what follows, we summarize some of the notable outcomes from the feasibility testing, highlighting with examples as space allows.\footnote{The Appendix details each prompt and response and provides a short analysis for each individual test condition.}

\textbf{Sensitivity of Models:} Generally, the content of GPT3.5 results was comparable to those from GPT4. However, GPT4 clearly excelled in generating formal descriptions of the problem space. Table~\ref{tab:compare-gpts} compares results for $A(4,9)\rightarrow 6$ for refined problem characterization. GPT3.5 results are comparatively less precise in the use of notation (e.g., there is no mapping from initial state values to specific pails) and also incorrect (the initial state would be an illegal state with its definition).  Because formal notation will be beneficial for future code generation, we focus on GPT4 outcomes for the remainder of the feasibility results. For the more fine-grained prompt/response interactions we envision in Figure~\ref{fig:cta-agent}, GPT3.5 (or other models of comparable scale) could still prove adequate in the long term.

\begin{table}[t]
\caption{Comparing GPT3.5 and GPT4 refined problem characterization results ($A(4,9)\rightarrow 5$).}
    \centering
    \begin{small}
    \begin{tabular}{p{.5in}p{5.15in}}
    \hline
         \textbf{GPT3.5} & 
Initial State: (0, 0) \par
Final State: (x, y) where x + y = 6 \par
Illegal/Impossible States: Any state where x + y $\neq$ 6 or x or y is negative. \\ \hline
         \textbf{GPT4} & 
Initial State (S0): \( (FHD1 = 0, FHD2 = 0) \), where \(FHD1\) and \(FHD2\) represent the amount of flucotone in each Flucotone-Holding Device respectively.\par
Goal State (SG): A state where the total amount of flucotone delivered equals 6 units. This can be represented as any state \( (FHD1, FHD2) \) where \( FHD1 + FHD2 = 6 \) and \( 0 \leq FHD1 \leq 4 \), \( 0 \leq FHD2 \leq 9 \). \par
Illegal States: Any state where \( FHD1 > 4 \) or \( FHD2 > 9 \)
\\ \hline
    \end{tabular}
      \end{small}
    
    \label{tab:compare-gpts}
\end{table}

\textbf{Familiar Cases:} 
As expected, a one-shot, chain-of-thought (CoT) prompt generally was sufficient to achieve acceptable problem-space formulations for the familiar cases.
For these cases, the only advantage of the CTA~Agent over one-shot CoT was that it enumerated better search control knowledge. For the familiar cases, a one-shot, CoT solution prompt provided correct solutions for $F(3,5)\rightarrow 4$ and sometimes $F(4,9)\rightarrow 6$ and even the $F(9,17)\rightarrow 5$ problem (generating a 20-step solution).

\textbf{Mixed-unit Variation ($V(4qt,9g)\rightarrow 6g$):} This variation demonstrated a significant improvement of the CTA Agent over the one-shot, CoT formulation, which never acknowledges or references the mixed units. The CoT formulation also has no quantity specification for the transfer operation, making it insufficient to use for an executable agent specification. The CTA Agent, in contrast, defines the initial state as (0,0) where ``the first element of the tuple represents the volume in gallons in the 4-quart pail: Note 4 quarts = 1 gallon.'' Given the correct definition of the state, the overall problem-space specification is sufficiently detailed for successful execution in a Soar agent. 

Finally, the CoT solver correctly identified the mixed units (``the 4-quart pail is essentially a 1-gallon pail'') but was not able to apply this information in subsequent steps (thus treating the 4-quart pail as a 4-gallon pail in its problem-solving steps).

\textbf{Extra Jug Variation ($V(2,3,5)\rightarrow 4$): } Similar to the mixed-unit variation, the CTA Agent formulation for this variation is complete and correct and sufficient for successful execution. It correctly specifies three values for the initial state (after refinement) and introduces two variables to represent which jugs to use in each transfer operation. The CoT formulation is incomplete (imprecise specification of transfer). The CoT solver generates the ``known'' solution to the \textit{Die Hard} problem, completely ignoring the 2-quart barrel and the more direct solution it affords.

\textbf{Analogue ($A:(4,9)\rightarrow 6$):} Again, the CTA Agent result is complete and correct and sufficient for execution by Soar. The one-shot CoT was incomplete (again, lacking detail in the transfer operators). However, the one-shot CoT solved the problem without issue, suggesting that this specific problem is a ``recognizable'' water-jugs analog to the LLM and thus insufficiently obscured.

\textbf{Refinement:} The reflection/refinement steps in the CTA~Agent were largely responsible for enabling sufficient and precise detail in the specification, differentiating the outcomes significantly compared to the one-shot CoT formulation. Table~\ref{tab:impact-of-refinement} illustrates representative examples. In the three-pail problem formulation, the initial state and goal states are inconsistent (4 quarts in the 2-quart pail). The revised formulation is much more precise. For operator specification, this example reflects a recurring pattern observed in the results: the introduction of an imprecise definition for the transfer operator is ``re-written'' in refinement to capture the specific conditionality of transfer.

\begin{table}[tb]
\caption{Role of refinement in improving problem-space formulation.}
    \centering
    \begin{small}
    \begin{tabular}{p{.5in}p{5.15in}}
    \hline
    \multicolumn{2}{l}{\textbf{Problem characterization ($V(2,3,5)\rightarrow 4$)}} \\
    Initial: & Initial State: (0, 0, 0) representing the amount of water in the 5-gallon, 3-gallon, and 2-gallon pails respectively.\par
Final State: (x, y, 4) where x and y can be any value such that the total amount of water in all containers equals 4 gallons. The primary goal is to have a total of 4 gallons, not necessarily in one container.\\
    \cline{2-2}
    Revised: & Initial State: (S$_{initial}$ = (0, 0, 0)), where each value represents the volume of water in the 5-gallon, 3-gallon, and 2-gallon containers respectively.\par

Goal State(s): \(S_{goal} = \{(x, y, z) | x + y + z = 4\}\), where \(x \leq 5\), \(y \leq 3\), and \(z \leq 2\). This set of states reflects all combinations where the total volume of water equals exactly 4 gallons, irrespective of the distribution across the containers.\\ \hline
        \multicolumn{2}{l}{\textbf{Operator specification for problem-space characterization} ($A(4,9)\rightarrow 6$)} \\
    Initial: & Transfer (FHDx, FHDy): \par Transfers flucotone from FHDx to FHDy until FHDx is empty or FHDy is full. \par
   - Preconditions: FHDx is not empty, and FHDy is not full. \par
   - Postconditions: FHDx decreases by the transferred amount, and FHDy increases by the same amount, without exceeding FHDy's capacity. \\ \cline{2-2}
    Revised: & Transfer(FHDx, FHDy) \par
   - Preconditions: (FHDx $>$ 0) AND (FHDy $<$ Capacity$_{FHDy}$) \par
   - Postconditions: (FHDx' = max(0, FHDx - (Capacity$_{FHDy}$ - FHDy))), (FHDy' = min(Capacity$_{FHDy}$, FHDy + FHDx)) \par
  Clarification on Transfer Operator: The original characterization did not explicitly account for the exact amount transferred. The revised postconditions now clearly define the new states of FHDx and FHDy after the transfer, ensuring that the transfer does not exceed FHDy's capacity or result in FHDx having a negative amount of flucotone.\\ \hline
    \end{tabular}
     \end{small}
    \label{tab:impact-of-refinement}
\end{table}

\textbf{Search Control:} The CTA Agent did identify useful search control guidelines. For example, for $V(2,3,5)\rightarrow 4$, the agent recognized all of the path constraints in the water-jobs problem-space formulation presented previously (Table~\ref{tab:probspace-example}). Search control is important: these simple rules of thumb in $F(9,17)\rightarrow 5$ decrease Soar's iterative deepening search over 2 orders of magnitude (over $2M$ decisions without search control to roughly $15K$ with these rules).

However, there are two clear gaps and requirements for future research. First, the LLM often asserted that non-monotonic changes to the contents of a container were undesired. However, most (interesting) water-jugs problems rely on the insight that discarding the contents of a container is required. Second, while the formal notation for states and operators was correct and precise, search control was more often expressed in language than notation. Even when notation was used, the notation was insufficient for direct translation to Soar. We can elaborate search control with refinement analysis, which will likely improve the results marginally. However, long term, determining how to encode search control may require tight integration with a specific problem-solving system (or the use of intermediate, declarative specification languages such as PDDL).

\textbf{Test Cases/Test States:} This step was the least successful of all the nodes we included in the CTA Agent. It produced test cases, but they were (generally) neither obvious examples of unit tests (e.g., representing the goal state to test goal recognition) nor alternative problem sets. Neither this step nor the problem characterization (which asked for examples of undesired states or configurations) identified problematic test cases (e.g., a problem with two pails of the same size). Again, in the long term, we can likely improve the generation of test cases, but the initial results from this node were limited at best.

\section{Conclusions}

We explored the feasibility of using agentic workflows with large language models to create specifications of problem spaces that are sufficient to enable knowledge-lean search for a solution in a problem-solving architecture like Soar. The results were strongly but not unequivocally positive. For the one class of problem we explored, the agentic approach provided sufficiently precise and correct problem-space specification to enable successful search in Soar. Further, in some cases, the agentic approach also identified search control knowledge that significantly decreases the resulting search space. The agentic roles defined in the Cognitive Task Analyst Agent were adequate for the feasibility evaluation, especially the role of an expert analyst critiquing and refining ongoing analysis in the \textit{Refinement} nodes in the agent architecture. 
Limitations of the feasibility results include 1) the focus on single family of problems (especially the question of the actual novelty of the analogous case), 2) less precise and useful contributions from the \textit{Search Control} and \textit{Test Case} generation nodes, and 3) the need for the larger scale LLM to generate sufficiently precise specifications for (future) code generation.

On balance, however, we conclude that this exploration does demonstrate the feasibility of a more full-featured CTA Agent. Next steps will focus on three enhancements to the multi-node/agent architecture described here as well as automated generation of PDDL (or similar). First, we will create many more nodes/agents, each with a more fine-grained task or goal than those described here. More fine-grained agents will result in more fine-grained, multi-agent interaction (e.g., between analyst and refine/QA roles). We expect such fine-grained interaction  will both improve results overall and have the potential to enable the use of smaller-scale LLMs (consistent with emerging uses of agentic workflows in other application domains). Second, we will extend the overall analysis architecture to a broader class of potential problem types (e.g., hill climbing and means-ends analysis). Multiple problem types will require developing a \textit{Problem-Solving Characteristics} agent (green node in Figure~\ref{fig:cta-agent}), which was not explored in the feasibility prototype. Third, we will also explore the definition of hierarchies of problem spaces rather than the simple problem-to-problem-space mapping assumed here. We also will evaluate with many more problem classes.

We hypothesize that the type of capability envisioned by the CTA Agent design could have a significant impact for the cognitive systems community. Completely automated, fully reliable problem-space specification may be a long-term effort. However, the success of the feasibility demonstration suggests that a significant fraction of knowledge creation for some types of cognitive-systems applications can be automated. A recurring challenge for this community is the requirement for human mediation of agent knowledge to enable research systems and applications. This mediation limits larger scientific impact because it is labor intensive to create broadly-capable and robust systems and also because researchers outside this community can be skeptical of the actual capability of the cognitive system relative to its developers. The automated problem specification approach has the potential to mitigate both of these issues and may also offer new directions for cognitive-systems research, such as alternative problem-representation strategies. Our expectation (and hope) is that such capability will contribute to faster and less mediated development of future cognitive systems.

\begin{acknowledgements} 
\noindent
This work was supported by the Office of Naval Research, contract N00014-21-1-2369. The views and conclusions contained in this document are those of the authors and should not be interpreted as representing the official policies, either expressed or implied, of the Department of Defense or Office of Naval Research. The U.S. Government is authorized to reproduce and distribute reprints for Government purposes notwithstanding any copyright notation hereon. We thank the anonymous reviewers for their constructive, detailed suggestions on improving the paper.\\
\end{acknowledgements} 

\vspace{-0.25in}

{\parindent -10pt\leftskip 10pt\noindent
\bibliographystyle{cogsysapa}
\bibliography{format,zotero-transoar}

}

\section*{Appendix}

This appendix summarizes prompts used in the CTA Agent and the response of nodes (LLM agents) to these prompts. The sections are:
\begin{itemize}
    \item CTA Agent Prompts (Including general system prompt and the instructional prompt for each node/agent)
    \item Chain-of-Thought Problem Space Formulation Prompt
        \item Chain-of-Thought Solution Prompt
    \item Results from Feasibility Test Cases
    \begin{itemize}
        \item Problem Description (used for all three experimental settings for each test case)
        \item LLM Response from CoT Solver
                \item LLM Response from CoT Problem-Space Formulation
        \item LLM Responses from CTA Agent (response from each individual node)        
    \end{itemize}
\end{itemize}

While both GPT3.5 and GPT4 models were explored in the development of the feasibility prototype, the results reported here are limited to GPT4 (model declarative from the source code below)

\begin{lstlisting}[breaklines]
model = ChatOpenAI(model="gpt-4-0125-preview", 
                   temperature=0, 
                   streaming=True)
\end{lstlisting}

\section*{Prompts for LLM Nodes in the CTA Agent}

This section summarizes each of the prompts used in the CTA Agent. The input prompt for each node/LLM agent in the system followed the format below. Braces denote variable substitution. The general systems prompt was the same for all agents (and also the CoT problem-formulation agent). The problem description was changed from test to test but was the same for each run of the agent. Each agent was then customized with specific instructions for the step in the analysis being for that node. The analysis instructions for each node are detailed in the remainder of this section.

\begin{lstlisting}[breaklines]
LLM_agent_template = """
{general_system_prompt}
SPECIFIC PROBLEM:
{problem_description}
{analysis_instructions}
LIMIT YOUR RESPONSE TO THIS QUESTION ONLY/ASPECT OF ANALYSIS."""
\end{lstlisting}

\subsection*{General Systems Prompt}
\nopagebreak
\begin{lstlisting}[breaklines]
You are an expert in cognitive task analysis. \
You are helping to design a reasoner/problem solver that can solve many different instances of a class of problem. \
Problems are defined in 1-2 paragraphs. There will be a specific example problem provided. \
However,  your responses should focus creating a general formulation of the problem space, not a problem space specific to the example only. \
Focus each response on the most recent, specific question asked of you. \
The questions are designed to break the problem-space formulation into a set of discrete steps. \
Unless otherwise directed, be concise in each response (i.e., respond in one sentence or expression).

Use these definitions:
Problem Space: A problem space consists a set of symbolic structures (the states of the space) and a set of operators over the space. \
Each operator takes a state as input and produces a state as output (although there may be other inputs and outputs as well). \
The operators may be partial (i.e., not defined for all states). Sequences of operators define paths that thread their way through sequences of states. 
         
Problem: A problem in a problem space consists of a set of initial states, a set of goal states, and a set of path constraints. \
The problem is to find a path through the space that starts at the initial state, passes only along paths that satisfy the path constraints, and ends at any goal state. 
\end{lstlisting}

\subsection*{Analysis Instructions for Problem Characterization}
\nopagebreak
\begin{lstlisting}[breaklines]
For this response, focus on characterizing the problem itself. \
What is the initial state? \
What is the final state? \
Are there illegal / impossible states that are not allowed for this problem? If so, identify them.
Use formal notation and/or mathematical expressions if/as possible to characterize the problem. Be concise.
\end{lstlisting}

\subsection*{Analysis Instructions for Refinement of Problem Characterization}
\nopagebreak
\begin{lstlisting}[breaklines]
You are an expert in Cognitive Task Analysis and specialize in reviewing the work of other analysts. \
Given the PROBLEM DESCRIPTION and the previous response characterizing the problem, improve the prior problem characterization.\
Are elements incorrect? If so, correct them. \
Are elements missing? If so, add them. Focus on problem characterization only (specifications of states). We will ask about actions and path constraints later. \
Are elements poorly formed or ambigious? Replace qualitative terms such as 'less' or 'more' with specific expressions that reflect quantitative values. \
Respond with a revised problem characterization that reflects your analysis. \
Use formal notation and/or mathematical expressions to characterize the problem. \
\end{lstlisting}

\subsection*{Analysis Instructions for Operators/Problem-Space Characterization}
\nopagebreak
\begin{lstlisting}[breaklines]
For this response, focus on characterizing the problem space that could be used to solve various instances of the problem. \
What actions/operators are relevant?
What are the preconditions for each operator?
What are the effects or postconditions of each operaetor?
Use formal notation and/or mathematical expressions if/as possible to characterize the problem space. Be concise.
\end{lstlisting}

\subsection*{Analysis Instructions for Refinement of Problem-Space Characterization}
\nopagebreak
\begin{lstlisting}[breaklines]
You are an expert in Cognitive Task Analysis and specialize in reviewing the work of other analysts.\
Given the PROBLEM DESCRIPTION and the previous AI response characterizing the problem, your task is to improve the problem space characterization.\
Does the set of operators appear complete? If not, suggest additional operators.
Are the preconditions for each operator correct? If not, corect them. 
Are the postconditions for each operator correct? If not, corect them. 
Focus on problem space characterization only (specifications of operators) rather than the state specification.
Respond with a revised problem space characterization (operators with pre- and post-conditions) that reflects your analysis. \
Use formal notation and/or mathematical expressions to characterize the problem space.
\end{lstlisting}

\subsection*{Analysis Instructions for Defining Search Control}
\nopagebreak
\begin{lstlisting}[breaklines]
The solution to this problem will be solved by a search from the initial state to the goal state using the operators specified thus far.\
Search can be made more efficient by identfying unproductive paths such as loops and dead-ends. Your task is to identify/specify these. 
What are undesirable states? For example, generating a state that is identical to the initial state is often not a productive step.
What are undesirable sequences of operator applications? For example, an operator sequence that undoes the action of the immediately previous operator is often not desirable.\
Use formal notation and/or mathematical expressions to characterize any undesirable states or operator sequences.
\end{lstlisting}

\subsection*{Analysis Instructions for Identifying Test States and Test Cases}
\nopagebreak
\begin{lstlisting}[breaklines]
You are an expert in Cognitive Task Analysis and specialize in identifying new problem instances.\
Given the PROBLEM DESCRIPTION and the previous AI responses characterizing the problem and problem space, \
your task is to identify specific cases that test/evaluate the problem space.\
Create three use cases. Number each case. For each case, specify the initial state and goal state using the notation from the problem characterization 
and describe (in one short sentence) why this is an apt test case for the problem space.\
Emphasize the generation of "edge" cases that test assumptions in the problem space.
\end{lstlisting}

\section*{Prompt for One-shot, Chain-of-Thought Solver}
\noindent Prompt:
\begin{lstlisting}[breaklines]
You are an expert problem solver with years of experience solving many different types of puzzles. What is your solution for this problem?

SPECIFIC PROBLEM:
{problem_description}

Think step-by-step in producing a response.
\end{lstlisting}

\section*{Prompt for One-shot, Chain-of-Thought Problem-Space Formulation}
This agent used the general system prompt above, followed by the problem description, and then the instruction prompt below, which is specific to only this condition. \\

\noindent Instruction Prompt:
\begin{lstlisting}[breaklines]
Produce a problem space formulation for the general problem, specifying initial and goal states, operators, \
and preconditions and postconditions for each operator. Identify any illegal or undesirable states. Introduce and use notation for describing states and operators.
\end{lstlisting}

\section*{Soar-Agent Solutions}
An existing Soar problem-solving agent was used to generate water-jug solutions for the test cases.\footnote{The agent is distributed with Soar as part of the Soar tutorial.} This agent uses iterative deepening, guaranteeing that an optimal (minimum step) solution is found. The agent uses Soar's impasse function to recognize that it does not what steps to apply to solve a (new) problem. It then generates new states in the search space by randomly choosing and applying operators in the current search state and then evaluating those new states.

Evaluations include identifying the goal state (search complete), new state (continue searching), depth criterion met (not the goal state but at the current iteration depth so return failure at this depth), or a failure state (see below). When Soar's chunking learning mechanism is enabled (``during learning'' and ``after learning'' in the results below), the results from evaluations are cached as the search proceeds. Thus, when a specific failure state is encountered after having seen it once, the current search path can be immediately abandoned. Learning to recognize failure states in this way can significantly reduce the total number of search states explored for a new problem. While such knowledge compilation is a hallmark of Soar systems \cite{laird_chunking_1986,laird_soar_2012,assanie_learning_2022}, it offers potential within the CTA Agent to offset the cost of incomplete generation of search control knowledge.

Tables~\ref{tab:app-soar-dec} and \ref{tab:app-search-states} summarize the number of Soar decisions (deliberate problem-solving steps) and the number of search states explored for the water-jug cases discussed in the paper. The second column in Table~\ref{tab:app-soar-dec} shows the minimum path length (number of problem-space operators) for each problem.  Note that the $A(4,9)\rightarrow6$ case is not included in the tables because, in terms of the Soar agent, results would be identical to the familiar $F(4,9)\rightarrow6$ case already listed.

We report results for a number of different test conditions for each case. First, we ran the agent with and without failure detection knowledge (the two primary columns in each table). For iterative deepening, we define failure states to be ones in which the jugs are all  empty or all full. This approach detects loops without requiring more detailed sequence tracking, reflecting a more sophisticated understanding of the problem (and mitigating the bookkeeping needing to avoid attempting to record all unnecessary search states). With failure detection, the agent evaluates all empty or all full states as dead ends and abandons them. In the agent without failure detection,  all full/all empty states are not marked as failures. As is evident in the results in Table~\ref{tab:app-search-states}, not abandoning these paths increases the effective branching factor of the resulting search space. By definition, with iterative deepening, the agent continually explores paths that include these states even thought these states are necessarily not on the (minimum-length) solution path that the iterative-deepening search eventually finds.

\begin{table}[htb]
\caption{Total Soar Agent Decisions for Various Agent Configurations for the Feasibility Evaluation Test Cases.}
    \centering
    \begin{tabular}{|l|r|rrr|rrr|}
    \hline
    & & \multicolumn{3}{c|}{with Failure Detection} & \multicolumn{3}{c|}{without Failure Detection} \\
    Test Case & \begin{turn}{75}Min. Soln \end{turn} & \begin{turn}{75} No Learning \end{turn}  & \begin{turn}{75} During Learning  \end{turn} & \begin{turn}{75} After Learning  \end{turn} & \begin{turn}{75} No Learning  \end{turn} & \begin{turn}{75} During Learning  \end{turn} & \begin{turn}{75} After Learning  \end{turn}\\ \hline

    $F(4,9)\rightarrow 6$ & 8 & 20,657 & 1357  & 9 & 80,434 & 2126 & 9\\ 
    $F(3,5)\rightarrow 4$ & 6 & 3355  & 548 & 7 & 8072 & 1087 & 7\\ 
    $F(9,17)\rightarrow 5$ & 20 & 86.9M & 8201 & 21 & \shortstack{600M+ \\ (see note)}& 7147 & 21 \\ 
    $F(4q,9g)\rightarrow 6g$ & 6 & 3422 & 526 & 7 & 8071 & 801 & 7  \\ 
    $F(2,3,5)\rightarrow 4$ & 4 &  1599 & 908 & 5 &1890 & 860 & 5\\     
         \hline
    \end{tabular}
    
    \label{tab:app-soar-dec}
\end{table}

\begin{table}[htb]
\caption{Search States Explored for Various Agent Configurations for the Feasibility Evaluation Test Cases.}
    \centering
    \begin{tabular}{|l|rrr|rrr|}
    \hline
    & \multicolumn{3}{c|}{with Failure Detection} & \multicolumn{3}{c|}{without Failure Detection} \\
    Test Case & \multicolumn{1}{l}{\begin{turn}{75} No Learning \end{turn}}  & \begin{turn}{75} During Learning  \end{turn} & \begin{turn}{75} After Learning  \end{turn} & \begin{turn}{75} No Learning  \end{turn} & \begin{turn}{75} During Learning  \end{turn} & \begin{turn}{75} After Learning  \end{turn}\\ \hline

    $F(4,9)\rightarrow 6$ &  5104 & 332 & 0 & 19055 & 556 & 0\\ 
    $F(3,5)\rightarrow 4$ & 841 & 89 & 0 & 1916 & 193 & 0 \\ 
    $F(9,17)\rightarrow 5$ & 21.4M & 1254 & 0 & (see note) & 1114 & 0 \\ 
    $F(4q,9g)\rightarrow 6g$ & 862 & 97 & 0  & 1929 & 136  & 0 \\ 
    $F(2,3,5)\rightarrow 4$ & 363& 212 & 0 & 421 & 199 & 0\\     
         \hline
    \end{tabular}
    
    \label{tab:app-search-states}
\end{table}

Within each of the two primary conditions (the two groups of three columns), we report and contrast three run-time settings: no learning, during learning, and after learning. For all configurations other than the $F(9,17)\rightarrow5$ no learning cases, the table shows the results averaged over 5 runs.\footnote{Results for $F(9,17)\rightarrow5$ for the ``no learning'' condition were just generated once, given the size of the search spaces required. Additionally, the operating system ran out of application memory in the terminal process after 600M decisions due to updates printed statements to the terminal process. We did not attempt to complete this run.}

These results demonstrate a few points relative to our overall investigation:

\begin{itemize}
    \item Problem solving reliability: Solutions to all problems are readily found by Soar, even for the problems that require millions of search states (for no learning configurations). This outcome is not surprising, but contrasts with LLM-only solutions, where ensuring a correct, reliable (and even repeatable) result is not guaranteed.
    \item Online learning has a large impact on required computation: Comparing the during-learning and no-learning cases to one another, one can observe the important impact of learning during search itself. In the case, the agent is compiling its evaluations of search states as it encounters them and is then able to recognize which states will lead to failure (including loops in the with failure detection case). This compilation has a huge impact on how many search states are needed as the number of steps in the solution increases.
    \item Importance of search control knowledge: Uncovering and including search control knowledge is sometimes not \textit{necessary} for a solution, but its absence can significantly increase the amount of computation needed to solve a problem. These results are most apparent in comparing the ``with'' and ``without failure detection'' no-learning cases. However, even when learning, comparing these two conditions, it is clear that learning can be used to cut off unproductive search paths much sooner and thus find solutions more quickly. The combination of search control and learning makes open-ended problem search much more feasible for even problems with a large branching factor for search.
    
\end{itemize}

\clearpage
\section*{Feasibility Evaluation Test Cases}
The following sections summarize results for each of test cases. Each section includes 1) the problem description provided to the CTA Agent nodes, 2) the results from one-shot, chain-of-thought solution prompt, 3) the results from one-shot, chain-of-thought problem-space-formulation prompt, and results from the CTA Agent.

\section*{Familiar: $F(4,9)\rightarrow6$}

\subsection*{$F(4,9)\rightarrow6$ Problem Description}

The general problem is to deliver a specific amount of water using containers that are opaque and have contain no graduated markings. \
The amount of water that needs to be delivered will generally differ from the full capacities of the container. As a specific example, \
how can you bring up from the river exactly 6 quarts of water when you have only two containers, \
a four quart pail and a nine quart pail to measure with?

\subsection*{Soar Agent Solution}
\begin{lstlisting}[breaklines]
     0: ==>S: S1
     1: O: O1 (initialize-water-jug-look-ahead)
S1: 4: 0
S1: 9: 0
     2: O: O3 (fill)
S1:  FILL(9)
S1: 9: 9
     3: O: O4 (pour)
S1:  POUR(9:9,4:0)
S1: 4: 4
S1: 9: 5
     4: O: O6 (empty)
S1:  EMPTY(4)
S1: 4: 0
     5: O: O12 (pour)
S1:  POUR(9:5,4:0)
S1: 4: 4
S1: 9: 1
     6: O: O14 (empty)
S1:  EMPTY(4)
S1: 4: 0
     7: O: O20 (pour)
S1:  POUR(9:1,4:0)
S1: 4: 1
S1: 9: 0
     8: O: O24 (fill)
S1:  FILL(9)
S1: 9: 9
     9: O: O26 (pour)
S1:  POUR(9:9,4:1)
S1: 4: 4
S1: 9: 6
S1 Solution Found!
\end{lstlisting}

\subsection*{One-shot, Chain-of-Thought Solution: $F(4,9)\rightarrow6$}
This solution is not correct. See Step 4.

\begin{lstlisting}[breaklines]
To solve this problem, we can use a method called the Water Jug Problem algorithm. Here's how you can bring up exactly 6 quarts of water using a 4-quart pail and a 9-quart pail:

1. Fill the 9-quart pail to its full capacity.
2. Pour the water from the 9-quart pail into the 4-quart pail until the 4-quart pail is full. This leaves 5 quarts of water in the 9-quart pail.
3. Empty the 4-quart pail.
4. Pour the remaining 5 quarts of water from the 9-quart pail into the 4-quart pail.
5. Fill the 9-quart pail to its full capacity again.
6. Pour water from the 9-quart pail into the 4-quart pail until the 4-quart pail is full. Since there are already 5 quarts in the 4-quart pail, you only need to pour 1 more quart to reach the desired 6 quarts.

By following these steps, you can accurately measure and deliver exactly 6 quarts of water using the 4-quart and 9-quart pails.
\end{lstlisting}

\subsection*{One-shot, Chain-of-Thought Problem-Space Formulation: $F(4,9)\rightarrow6$}
This formulation is generally correct and reasonably formal. The search control is not helpful (and expressed largely in language).

\begin{lstlisting}[breaklines]
### Problem Space Formulation for the General Problem

#### States
- A state can be represented as a tuple \((C_1, C_2, \ldots, C_n, T)\), where \(C_i\) represents the current amount of water in container \(i\) (for \(i = 1\) to \(n\), where \(n\) is the number of containers), and \(T\) represents the target amount of water to be delivered. 
- Initial State: \((0, 0, \ldots, 0, T)\), assuming all containers are initially empty and \(T\) is the target amount of water.
- Goal State: Any state \((C_1, C_2, \ldots, C_n, T)\) where at least one \(C_i = T\) or the sum of any combination of \(C_i\) equals \(T\).

#### Operators
1. **Fill**: Completely fill a container from the water source.
   - **Precondition**: Container \(i\) is not full.
   - **Postcondition**: Container \(i\) is full, \(C_i = \text{Capacity of } i\).
2. **Empty**: Empty a container.
   - **Precondition**: Container \(i\) is not empty.
   - **Postcondition**: Container \(i\) is empty, \(C_i = 0\).
3. **Transfer**: Transfer water from one container to another until the source container is empty or the destination container is full.
   - **Precondition**: Container \(i\) is not empty, and container \(j\) is not full.
   - **Postcondition**: Water is transferred from \(i\) to \(j\) until \(i\) is empty or \(j\) is full, updating \(C_i\) and \(C_j\) accordingly.

#### Preconditions and Postconditions
- **Fill**:
  - Pre: \(C_i < \text{Capacity of } i\)
  - Post: \(C_i = \text{Capacity of } i\)
- **Empty**:
  - Pre: \(C_i > 0\)
  - Post: \(C_i = 0\)
- **Transfer**:
  - Pre: \(C_i > 0\) and \(C_j < \text{Capacity of } j\)
  - Post: \(C_i' = \max(0, C_i - (\text{Capacity of } j - C_j))\) and \(C_j' = \min(\text{Capacity of } j, C_j + C_i)\)

#### Illegal or Undesirable States
- Any state where \(C_i > \text{Capacity of } i\) for any container \(i\) is illegal (overfilling a container).
- Any state where the sum of water in all containers exceeds the target amount without any single container or combination of containers exactly meeting the target amount is undesirable, as it does not solve the problem.
\end{lstlisting}

\subsection*{CTA Agent: $F(4,9)\rightarrow6$}
This formulation is complete and precise. Interestingly (compared to many of the other results), these refinement/revision results are more informal/``chatty'' and do not use notation as reliably as some of the other tests. The description and semi-formalization of loop detection is excellent for this case.

\subsubsection*{Input Prompt}
\begin{lstlisting}[breaklines]

You are an expert in cognitive task analysis. You are helping to design a reasoner/problem solver that can solve many different instances of a class of problem. Problems are defined in 1-2 paragraphs. There will be a specific example problem provided. However,  your responses should focus creating a general formulation of the problem space, not a problem space specific to the example only. Focus each response on the most recent, specific question asked of you. The questions are designed to break the problem-space formulation into a set of discrete steps. Unless otherwise directed, be concise in each response (i.e., respond in one sentence or expression).

Use these definitions:
Problem Space: A problem space consists a set of symbolic structures (the states of the space) and a set of operators over the space. Each operator takes a state as input and produces a state as output (although there may be other inputs and outputs as well). The operators may be partial (i.e., not defined for all states). Sequences of operators define paths that thread their way through sequences of states. 
         
Problem: A problem in a problem space consists of a set of initial states, a set of goal states, and a set of path constraints. The problem is to find a path through the space that starts at the initial state, passes only along paths that satisfy the path constraints, and ends at any goal state. 

PROBLEM DESCRIPTION:

The general problem is to deliver a specific amount of water using containers that are opaque and have contain no graduated markings. The amount of water that needs to be delivered will generally differ from the full capacities of the container. As a specific example, how can you bring up from the river exactly 6 quarts of water when you have only two containers, a four quart pail and a nine quart pail to measure with?

The solution to this problem will be solved by a search from the initial state to the goal state using the operators specified thus far.Search can be made more efficient by identfying unproductive paths such as loops and dead-ends. Your task is to identify/specify these. 
What are undesirable states? For example, generating a state that is identical to the initial state is often not a productive step.
What are undesirable sequences of operator applications? For example, an operator sequence that undoes the action of the immediately previous operator is often not desirable.Use formal notation and/or mathematical expressions to characterize any undesirable states or operator sequences.


LIMIT YOUR RESPONSE TO JUST THIS QUESTION/ASPECT OF ANALYSIS.
\end{lstlisting}

\subsubsection*{Initial Problem Characterization}
\begin{lstlisting}[breaklines]
Initial State: Both containers are empty, represented as (0, 0), where the first value in the tuple represents the amount of water in the four-quart pail and the second value represents the amount of water in the nine-quart pail.

Final State: The goal is to have exactly 6 quarts of water in one of the containers, which can be represented as either (4, 6) or (6, 3) or any state where the total amount of water equals 6 quarts, with the understanding that the excess water, if any, must be in the other container or discarded.

Illegal/Impossible States: Any state where the amount of water in either container exceeds its capacity or is negative, i.e., states where the first value is not in the range [0, 4] or the second value is not in the range [0, 9], are considered illegal or impossible states.
\end{lstlisting}

\subsubsection*{Refinement of Problem Characterization}
\begin{lstlisting}[breaklines]
The revised problem characterization, focusing on precision and clarity, is as follows:

Initial State: \(S_{initial} = (0, 0)\), where the first element of the tuple represents the volume of water in the four-quart pail, and the second element represents the volume of water in the nine-quart pail.

Goal State(s): \(S_{goal} = \{(x, y) | x + y = 6\}\) and \(x \leq 4\) and \(y \leq 9\), indicating that the sum of the volumes in both containers equals 6 quarts, within the constraints of their capacities.

Illegal States: \(S_{illegal} = \{(x, y) | x > 4 \lor y > 9 \lor x < 0 \lor y < 0\}\), representing any state where the volume in either container exceeds its capacity or is negative.
\end{lstlisting}

\subsubsection*{Initial Operator/Problem-Space Characterization}
\begin{lstlisting}[breaklines]
Apologies for the confusion in my previous responses. Let's accurately define the actions/operators, preconditions, and postconditions for the problem space:

### Operators
1. **Fill(p)**: Completely fill pail \(p\) from the river.
   - Preconditions: \(p\) is not full.
   - Postconditions: \(p\) is full.
   
2. **Empty(p)**: Empty all the water from pail \(p\).
   - Preconditions: \(p\) contains some water.
   - Postconditions: \(p\) is empty.
   
3. **Transfer(p1, p2)**: Pour water from pail \(p1\) to pail \(p2\) until \(p1\) is empty or \(p2\) is full.
   - Preconditions: \(p1\) is not empty and \(p2\) is not full.
   - Postconditions: Water volume transferred from \(p1\) to \(p2\) until \(p1\) is empty or \(p2\) is full.

### Formal Notation
Let \(p_4\) and \(p_9\) represent the four-quart and nine-quart pails, respectively. The state can be represented as a tuple \((v_4, v_9)\), where \(v_4\) and \(v_9\) are the current volumes of water in \(p_4\) and \(p_9\), respectively.

1. **Fill(p)**:
   - Preconditions: \(v_p < \text{Capacity}(p)\)
   - Postconditions: \(v_p' = \text{Capacity}(p)\)
   
2. **Empty(p)**:
   - Preconditions: \(v_p > 0\)
   - Postconditions: \(v_p' = 0\)
   
3. **Transfer(p1, p2)**:
   - Preconditions: \(v_{p1} > 0 \land v_{p2} < \text{Capacity}(p2)\)
   - Postconditions: \(v_{p1}' = \max(0, v_{p1} + v_{p2} - \text{Capacity}(p2))\), \(v_{p2}' = \min(\text{Capacity}(p2), v_{p1} + v_{p2})\)

These operators, along with their preconditions and postconditions, define the actions that can be taken within the problem space to move from the initial state towards the goal state(s).
\end{lstlisting}

\subsubsection*{Refinement of Operator/Problem-Space Characterization}
\begin{lstlisting}[breaklines]
Upon further review, it's clear that the initial response did not fully address the question regarding the completeness and correctness of the set of operators, including their preconditions and postconditions. Here is a refined and corrected version of the problem space characterization:

### Operators for the Water Jug Problem

1. **Fill(p)**: Fill pail \(p\) to its full capacity.
   - Preconditions: \(p\) is not already full.
   - Postconditions: \(p\) becomes full, i.e., if \(p\) is the four-quart pail, its volume becomes 4 quarts; if \(p\) is the nine-quart pail, its volume becomes 9 quarts.

2. **Empty(p)**: Empty pail \(p\).
   - Preconditions: \(p\) contains some water (volume > 0).
   - Postconditions: \(p\) becomes empty, i.e., its volume becomes 0.

3. **Transfer(p1, p2)**: Transfer water from pail \(p1\) to pail \(p2\) until \(p1\) is empty or \(p2\) is full, whichever comes first.
   - Preconditions: \(p1\) contains water (volume > 0) and \(p2\) is not full.
   - Postconditions: Water is transferred from \(p1\) to \(p2\) until \(p1\) is empty or \(p2\) is full. The volumes of \(p1\) and \(p2\) are adjusted accordingly.

### Corrections and Additions

- **Addition of a Measure Operator**: It seems necessary to explicitly define an operator for measuring the water if the goal involves having an exact amount of water that does not match the containers' capacities. However, given the problem constraints, such an operator cannot be directly applied because the containers are opaque and unmarked. Therefore, the solution relies on the strategic use of fill, empty, and transfer operations to achieve the goal state.

- **Clarification on Preconditions and Postconditions**:
  - For **Fill(p)**, the precondition should explicitly state that the pail's current volume is less than its capacity. The postcondition should specify the exact volume after the operation.
  - For **Empty(p)**, the precondition is correctly stated as the pail containing some water. The postcondition correctly sets the pail's volume to 0.
  - For **Transfer(p1, p2)**, the preconditions and postconditions need to account for the volumes before and after the transfer, ensuring that the sum of the volumes in \(p1\) and \(p2\) remains constant, and neither pail exceeds its capacity nor goes below 0 quarts.

### Formal Notation Correction

- The formal notation provided accurately captures the essence of the operators but could be enhanced by specifying the mathematical expressions for the volume changes during the transfer operation, ensuring clarity in the postconditions for all possible outcomes of the transfer.

In summary, the initial set of operators is fundamentally sound but requires minor adjustments and clarifications, particularly in the precise definition of postconditions for the transfer operation. The addition of an explicit measuring or calculation step is not feasible within the given constraints, emphasizing the need for strategic application of the defined operators to achieve the goal state.
\end{lstlisting}

\subsubsection*{Generation of Search Control}
\begin{lstlisting}[breaklines]
Undesirable states in this problem are those where the actions taken do not progress towards achieving the goal state of having exactly 6 quarts of water in any container. Specifically:

1. **States with the same volume of water as previously encountered states**, which indicates a loop. For example, returning to state (0, 0) after any sequence of operations.

2. **States where the total volume of water does not change towards achieving the goal state**, such as transferring water between containers without getting closer to having a total of 6 quarts in one of them.

Undesirable sequences of operator applications include:

1. **Applying an operator that undoes the immediately previous operator's action**, such as:
   - Filling a pail and then immediately emptying it without any intermediate beneficial action, represented as \(Fill(p) \rightarrow Empty(p)\) or \(Empty(p) \rightarrow Fill(p)\) without changing the state of the other pail in between.
   - Transferring water from one pail to another and then immediately transferring it back without any intermediate action that changes the state towards the goal, represented as \(Transfer(p1, p2) \rightarrow Transfer(p2, p1)\) without any beneficial change in water volume towards achieving the goal.

2. **Repeatedly filling and emptying the same pail without making progress towards the goal**, which can be seen as a specific case of the first undesirable sequence but is worth highlighting due to its potential frequency in naive solution attempts.

Formally, these can be characterized as follows:

- **Loop Detection**: If a state \(S_{current} = (v_4, v_9)\) is identical to any previously encountered state \(S_{previous}\), then the sequence of operations leading back to \(S_{current}\) is undesirable.

- **Unproductive Operator Sequences**: If an operation \(O_1\) directly followed by \(O_2\) results in a state \(S'\) that is either identical to the initial state \(S\) or does not bring the system closer to the goal state in terms of the total volume of water being closer to 6 quarts, then the sequence \([O_1, O_2]\) is undesirable. This can be mathematically represented as:
  - For \(Fill(p) \rightarrow Empty(p)\) or \(Empty(p) \rightarrow Fill(p)\), if \(S' = S\), then the sequence is undesirable.
  - For \(Transfer(p1, p2) \rightarrow Transfer(p2, p1)\), if \(S' = S\) or \(|6 - (v_4' + v_9')| \geq |6 - (v_4 + v_9)|\), then the sequence is undesirable, where \(v_4', v_9'\) are the volumes of water in the pails after the sequence of operations.
\end{lstlisting}

\clearpage
\section*{Familiar: $F(3,5)\rightarrow4$}

\subsection*{$F(3,5)\rightarrow4$ Problem Description}

The general problem is to deliver a specific amount of water using containers that are opaque and have contain no graduated markings. \
The amount of water that needs to be delivered will generally differ from the full capacities of the container. As a specific example, \
how can you bring up from the river exactly 4 gallons of water when you have only two containers, \
a five gallon pail and a three gallon pail to measure with?

\subsection*{Soar Agent Solution}

\begin{lstlisting}[breaklines]
    0: ==>S: S1
     1: O: O1 (initialize-water-jug-look-ahead)
S1: 3: 0
S1: 5: 0
     2: O: O3 (fill)
S1:  FILL(5)
S1: 5: 5
     3: O: O4 (pour)
S1:  POUR(5:5,3:0)
S1: 3: 3
S1: 5: 2
     4: O: O6 (empty)
S1:  EMPTY(3)
S1: 3: 0
     5: O: O12 (pour)
S1:  POUR(5:2,3:0)
S1: 3: 2
S1: 5: 0
     6: O: O16 (fill)
S1:  FILL(5)
S1: 5: 5
     7: O: O18 (pour)
S1:  POUR(5:5,3:2)
S1: 3: 3
S1: 5: 4
S1 Solution Found!
\end{lstlisting}

\subsection*{One-shot, Chain-of-Thought Solution: $F(3,5)\rightarrow4$}
This solution is correct.
\begin{lstlisting}[breaklines]
To solve this problem, we can follow these steps:

1. Fill the 5-gallon pail completely with water from the river.
2. Pour the water from the 5-gallon pail into the 3-gallon pail until the 3-gallon pail is full. This leaves 2 gallons of water in the 5-gallon pail.
3. Empty the 3-gallon pail.
4. Pour the remaining 2 gallons of water from the 5-gallon pail into the empty 3-gallon pail.
5. Fill the 5-gallon pail again.
6. Pour water from the 5-gallon pail into the 3-gallon pail until the 3-gallon pail is full. This will take 1 gallon of water from the 5-gallon pail, leaving exactly 4 gallons of water in the 5-gallon pail.

By following these steps, you can accurately measure and deliver exactly 4 gallons of water using the 5-gallon and 3-gallon pails.
\end{lstlisting}

\subsection*{One-shot, Chain-of-Thought Problem-Space Formulation: $F(3,5)\rightarrow4$}
This formulation is correct and the operator definitions are sufficiently precise for execution.  Search control outputs are not particularly useful.

\begin{lstlisting}[breaklines]
### Problem Space Formulation for the General Problem

#### States
- A state can be represented as a tuple \((C_1, C_2, \ldots, C_n, T)\), where \(C_i\) represents the current amount of water in container \(i\) (for \(i = 1\) to \(n\), where \(n\) is the number of containers), and \(T\) represents the target amount of water to be delivered. 
- Initial State: \((0, 0, \ldots, 0, T)\), assuming all containers are initially empty and \(T\) is the target amount of water.
- Goal State: Any state \((C_1, C_2, \ldots, C_n, T)\) where at least one \(C_i = T\) or the sum of any combination of \(C_i\) equals \(T\).

#### Operators
1. **Fill**: Completely fill a container from the water source.
   - **Precondition**: Container \(i\) is not full.
   - **Postcondition**: Container \(i\) is full, \(C_i = \text{Capacity of } i\).
2. **Empty**: Empty a container.
   - **Precondition**: Container \(i\) is not empty.
   - **Postcondition**: Container \(i\) is empty, \(C_i = 0\).
3. **Transfer**: Transfer water from one container to another until the source container is empty or the destination container is full.
   - **Precondition**: Container \(i\) is not empty, and container \(j\) is not full.
   - **Postcondition**: Water is transferred from \(i\) to \(j\) until \(i\) is empty or \(j\) is full, updating \(C_i\) and \(C_j\) accordingly.

#### Preconditions and Postconditions
- **Fill**:
  - Pre: \(C_i < \text{Capacity of } i\)
  - Post: \(C_i = \text{Capacity of } i\)
- **Empty**:
  - Pre: \(C_i > 0\)
  - Post: \(C_i = 0\)
- **Transfer**:
  - Pre: \(C_i > 0\) and \(C_j < \text{Capacity of } j\)
  - Post: \(C_i' = \max(0, C_i - (\text{Capacity of } j - C_j))\) and \(C_j' = \min(\text{Capacity of } j, C_j + C_i)\)

#### Illegal or Undesirable States
- Any state where \(C_i > \text{Capacity of } i\) for any container \(i\) is illegal (overfilling a container).
- Any state where the sum of water in all containers exceeds the target amount without any single container or combination of containers exactly meeting the target amount is undesirable, as it does not solve the problem.
\end{lstlisting}

\subsection*{CTA Agent: $F(3,5)\rightarrow4$}
This formulation is somewhat correct and sufficiently precise. This example is the well-known \textit{Die Hard} problem (see above; GPT volunteers that it recognizes this problem this way) and one potential consequence of that familiarity is that the operators are defined as specific to 3-quart and 5-quart volumes. This over-specificity did recur elsewhere in the data.

\subsubsection*{Input Prompt}
\begin{lstlisting}[breaklines]

You are an expert in cognitive task analysis. You are helping to design a reasoner/problem solver that can solve many different instances of a class of problem. Problems are defined in 1-2 paragraphs. There will be a specific example problem provided. However,  your responses should focus creating a general formulation of the problem space, not a problem space specific to the example only. Focus each response on the most recent, specific question asked of you. The questions are designed to break the problem-space formulation into a set of discrete steps. Unless otherwise directed, be concise in each response (i.e., respond in one sentence or expression).

Use these definitions:
Problem Space: A problem space consists a set of symbolic structures (the states of the space) and a set of operators over the space. Each operator takes a state as input and produces a state as output (although there may be other inputs and outputs as well). The operators may be partial (i.e., not defined for all states). Sequences of operators define paths that thread their way through sequences of states. 
         
Problem: A problem in a problem space consists of a set of initial states, a set of goal states, and a set of path constraints. The problem is to find a path through the space that starts at the initial state, passes only along paths that satisfy the path constraints, and ends at any goal state. 

PROBLEM DESCRIPTION:

The general problem is to deliver a specific amount of water using containers that are opaque and have contain no graduated markings. The amount of water that needs to be delivered will generally differ from the full capacities of the container. As a specific example, how can you bring up from the river exactly 4 gallons of water when you have only two containers, a five gallon pail and a three gallon pail to measure with?

The solution to this problem will be solved by a search from the initial state to the goal state using the operators specified thus far.Search can be made more efficient by identfying unproductive paths such as loops and dead-ends. Your task is to identify/specify these. 
What are undesirable states? For example, generating a state that is identical to the initial state is often not a productive step.
What are undesirable sequences of operator applications? For example, an operator sequence that undoes the action of the immediately previous operator is often not desirable.Use formal notation and/or mathematical expressions to characterize any undesirable states or operator sequences.


LIMIT YOUR RESPONSE TO JUST THIS QUESTION/ASPECT OF ANALYSIS.
\end{lstlisting}

\subsubsection*{Initial Problem Characterization}
\begin{lstlisting}[breaklines]
Initial State: (0, 0) where the first value represents the amount of water in the five-gallon pail and the second value represents the amount of water in the three-gallon pail.

Final State: (4, x) where x can be any value representing the amount of water in the three-gallon pail, as the problem only specifies needing exactly 4 gallons in the five-gallon pail.

Illegal/Impossible States: Any state where the amount of water in the five-gallon pail is greater than 5 or less than 0, and any state where the amount of water in the three-gallon pail is greater than 3 or less than 0. Formally, these are states (a, b) where a > 5 or a < 0, or b > 3 or b < 0.
\end{lstlisting}

\subsubsection*{Refinement of Problem Characterization}
\begin{lstlisting}[breaklines]
The revised problem characterization, focusing on precision and clarity, is as follows:

Initial State: \(S_{initial} = (0, 0)\), where the first element in the tuple represents the volume of water in the five-gallon pail (\(V_{5gal}\)), and the second element represents the volume of water in the three-gallon pail (\(V_{3gal}\)).

Goal State: \(S_{goal} = \{(4, y) | y \in \mathbb{Z}, 0 \leq y \leq 3\}\), indicating that the goal is achieved when the five-gallon pail contains exactly 4 gallons of water, and the three-gallon pail can contain any amount of water from 0 to 3 gallons inclusive.

Constraints on States: Given the containers' capacities, the set of all possible states is defined as \(S = \{(x, y) | x \in \mathbb{Z}, 0 \leq x \leq 5, y \in \mathbb{Z}, 0 \leq y \leq 3\}\). This formalization excludes any states where the volume of water in either pail exceeds its capacity or is negative, ensuring all states are physically possible given the problem constraints.
\end{lstlisting}

\subsubsection*{Initial Operator/Problem-Space Characterization}
\begin{lstlisting}[breaklines]
Operators and their characteristics for the problem space are as follows:

1. **Fill Operator (\(Fill_i\))**:
   - **Precondition**: Container \(i\) is not full.
   - **Action**: Fill container \(i\) to its maximum capacity.
   - **Effect**: \(S_{new} = (max_i, y)\) or \(S_{new} = (x, max_j)\), where \(max_i\) and \(max_j\) are the capacities of containers \(i\) and \(j\), respectively.

2. **Empty Operator (\(Empty_i\))**:
   - **Precondition**: Container \(i\) contains some water (\(> 0\)).
   - **Action**: Empty all water from container \(i\).
   - **Effect**: \(S_{new} = (0, y)\) or \(S_{new} = (x, 0)\), depending on which container is emptied.

3. **Transfer Operator (\(Transfer_{i \rightarrow j}\))**:
   - **Precondition**: Container \(i\) has water, and container \(j\) is not full.
   - **Action**: Transfer water from container \(i\) to container \(j\) until container \(i\) is empty or container \(j\) is full.
   - **Effect**: \(S_{new} = (x - \Delta, y + \Delta)\) or \(S_{new} = (x + \Delta, y - \Delta)\), where \(\Delta = \min(x, max_j - y)\) or \(\Delta = \min(y, max_i - x)\), ensuring the transfer does not exceed the capacity of the receiving container or the available water in the transferring container.

These operators define the actions available in the problem space, with their preconditions ensuring actions are only taken when possible and their effects updating the state based on the action taken.
\end{lstlisting}

\subsubsection*{Refinement of Operator/Problem-Space Characterization}
\begin{lstlisting}[breaklines]
The revised problem space characterization, focusing on operators with their preconditions and postconditions, is as follows:

1. **Fill Operator for Five-Gallon Pail (\(Fill_5\))**:
   - **Precondition**: \(V_{5gal} < 5\)
   - **Postcondition**: \(V_{5gal} = 5\)

2. **Fill Operator for Three-Gallon Pail (\(Fill_3\))**:
   - **Precondition**: \(V_{3gal} < 3\)
   - **Postcondition**: \(V_{3gal} = 3\)

3. **Empty Operator for Five-Gallon Pail (\(Empty_5\))**:
   - **Precondition**: \(V_{5gal} > 0\)
   - **Postcondition**: \(V_{5gal} = 0\)

4. **Empty Operator for Three-Gallon Pail (\(Empty_3\))**:
   - **Precondition**: \(V_{3gal} > 0\)
   - **Postcondition**: \(V_{3gal} = 0\)

5. **Transfer Operator from Five to Three (\(Transfer_{5\rightarrow3}\))**:
   - **Precondition**: \(V_{5gal} > 0\) and \(V_{3gal} < 3\)
   - **Postcondition**: \(V_{5gal} = V_{5gal} - \min(V_{5gal}, 3 - V_{3gal})\), \(V_{3gal} = V_{3gal} + \min(V_{5gal}, 3 - V_{3gal})\)

6. **Transfer Operator from Three to Five (\(Transfer_{3\rightarrow5}\))**:
   - **Precondition**: \(V_{3gal} > 0\) and \(V_{5gal} < 5\)
   - **Postcondition**: \(V_{3gal} = V_{3gal} - \min(V_{3gal}, 5 - V_{5gal})\), \(V_{5gal} = V_{5gal} + \min(V_{3gal}, 5 - V_{5gal})\)

These operators are designed to ensure that the actions taken are valid within the constraints of the problem, specifically the capacities of the containers and the goal of measuring out a specific amount of water. The preconditions prevent actions that are not possible given the current state (e.g., transferring water from an empty container), and the postconditions accurately reflect the state of the system following the action.
\end{lstlisting}

\subsubsection*{Generation of Search Control}
\begin{lstlisting}[breaklines]
Undesirable states and sequences in this problem space include:

1. **Undesirable States**:
   - States where the total volume of water does not change towards achieving the goal state, such as repeatedly filling and emptying the same container without transferring water to the other container. Formally, any sequence of operations that results in a state \(S = (x, y)\) that is revisited without any intermediate state \(S' = (4, y')\) being a goal state is undesirable.

2. **Undesirable Sequences of Operator Applications**:
   - Sequences that involve filling a container and then immediately emptying it without any intermediate beneficial action, e.g., \(Fill_i; Empty_i\) without a transfer operation in between.
   - Sequences that involve transferring water from one container to another and then immediately reversing the transfer, e.g., \(Transfer_{i \rightarrow j}; Transfer_{j \rightarrow i}\) without any intermediate operation that changes the state towards the goal.
   - Any sequence that leads to a loop, where the same set of states is visited repeatedly without progress towards the goal state, e.g., \(Fill_i; Transfer_{i \rightarrow j}; Empty_j; Transfer_{j \rightarrow i}\) repeated indefinitely.

Formally, undesirable sequences can be characterized as any sequence of operations \(\sigma\) where the application of \(\sigma\) on a state \(S\) results in a state \(S'\) such that \(S = S'\) or does not bring the system closer to any \(S_{goal}\), considering the defined goal states.
\end{lstlisting}

\clearpage
\section*{Familiar: $F(9,17)\rightarrow5$}

\subsection*{$F(9,17)\rightarrow5$ Problem Description}

The general problem is to deliver a specific amount of water using containers that are opaque and have contain no graduated markings. \
The amount of water that needs to be delivered will generally differ from the full capacities of the container. As a specific example, \
how can you bring up from the river exactly 5 gallons of water when you have only two containers, \
a 17-gallon pail and a 9-gallon pail to measure with?

\subsection*{Soar Agent Solution}

\begin{lstlisting}[breaklines]
      0: ==>S: S1 
     1:    O: O1 (initialize-water-jug-look-ahead)
S1: 17: 0
S1: 9: 0
     2:    O: O2 (fill)
S1:  FILL(17)
S1: 17: 17
     3:    O: O4 (pour)
S1:  POUR(17:17,9:0)
S1: 9: 9
S1: 17: 8
     4:    O: O6 (empty)
S1:  EMPTY(9)
S1: 9: 0
     5:    O: O12 (pour)
S1:  POUR(17:8,9:0)
S1: 9: 8
S1: 17: 0
     6:    O: O16 (fill)
S1:  FILL(17)
S1: 17: 17
     7:    O: O18 (pour)
S1:  POUR(17:17,9:8)
S1: 9: 9
S1: 17: 16
     8:    O: O20 (empty)
S1:  EMPTY(9)
S1: 9: 0
     9:    O: O26 (pour)
S1:  POUR(17:16,9:0)
S1: 9: 9
S1: 17: 7
    10:    O: O28 (empty)
S1:  EMPTY(9)
S1: 9: 0
    11:    O: O34 (pour)
S1:  POUR(17:7,9:0)
S1: 9: 7
S1: 17: 0
    12:    O: O38 (fill)
S1:  FILL(17)
S1: 17: 17
    13:    O: O40 (pour)
S1:  POUR(17:17,9:7)
S1: 9: 9
S1: 17: 15
    14:    O: O42 (empty)
S1:  EMPTY(9)
S1: 9: 0
    15:    O: O48 (pour)
S1:  POUR(17:15,9:0)
S1: 9: 9
S1: 17: 6
    16:    O: O50 (empty)
S1:  EMPTY(9)
S1: 9: 0
    17:    O: O56 (pour)
S1:  POUR(17:6,9:0)
S1: 9: 6
S1: 17: 0
    18:    O: O60 (fill)
S1:  FILL(17)
S1: 17: 17
    19:    O: O62 (pour)
S1:  POUR(17:17,9:6)
S1: 9: 9
S1: 17: 14
    20:    O: O64 (empty)
S1:  EMPTY(9)
S1: 9: 0
    21:    O: O70 (pour)
S1:  POUR(17:14,9:0)
S1: 9: 9
S1: 17: 5
S1 Solution Found!
\end{lstlisting}

\subsection*{One-shot, Chain-of-Thought Solution: $F(9,17)\rightarrow5$}
This 20-step solution is correct (although not the shortest path solution).

\begin{lstlisting}[breaklines]
To solve this problem, you can use the two containers (17-gallon and 9-gallon) to measure out exactly 5 gallons of water by following these steps:

1. **Fill the 17-gallon pail completely from the river.** You now have 17 gallons of water.

2. **Use the 17-gallon pail to fill the 9-gallon pail.** Pour water from the 17-gallon pail into the 9-gallon pail until the 9-gallon pail is full. You will be left with 17 - 9 = 8 gallons of water in the 17-gallon pail.

3. **Empty the 9-gallon pail back into the river.** Now, you have 8 gallons in the 17-gallon pail and an empty 9-gallon pail.

4. **Transfer the 8 gallons from the 17-gallon pail to the 9-gallon pail.** Now, the 17-gallon pail is empty, and the 9-gallon pail has 8 gallons of water.

5. **Fill the 17-gallon pail completely from the river again.** You now have 17 gallons in the 17-gallon pail and 8 gallons in the 9-gallon pail.

6. **Top off the 9-gallon pail using the 17-gallon pail.** Since the 9-gallon pail already has 8 gallons, it only needs 1 more gallon to be full. Pour water from the 17-gallon pail into the 9-gallon pail until it is full. This leaves you with 17 - 1 = 16 gallons of water in the 17-gallon pail.

7. **Empty the 9-gallon pail back into the river.** Now, you have 16 gallons in the 17-gallon pail.

8. **Fill the 9-gallon pail from the 17-gallon pail.** This leaves you with 16 - 9 = 7 gallons in the 17-gallon pail.

9. **Empty the 9-gallon pail back into the river again.**

10. **Transfer the 7 gallons from the 17-gallon pail to the 9-gallon pail.** Now, the 17-gallon pail is empty, and the 9-gallon pail has 7 gallons.

11. **Fill the 17-gallon pail completely from the river again.** You now have 17 gallons in the 17-gallon pail and 7 gallons in the 9-gallon pail.

12. **Top off the 9-gallon pail using the 17-gallon pail.** Since the 9-gallon pail already has 7 gallons, it only needs 2 more gallons to be full. Pour water from the 17-gallon pail into the 9-gallon pail until it is full. This leaves you with 17 - 2 = 15 gallons of water in the 17-gallon pail.

13. **Empty the 9-gallon pail back into the river.** Now, you have 15 gallons in the 17-gallon pail.

14. **Fill the 9-gallon pail from the 17-gallon pail.** This leaves you with 15 - 9 = 6 gallons in the 17-gallon pail.

15. **Empty the 9-gallon pail back into the river again.**

16. **Transfer the 6 gallons from the 17-gallon pail to the 9-gallon pail.** Now, the 17-gallon pail is empty, and the 9-gallon pail has 6 gallons.

17. **Fill the 17-gallon pail completely from the river again.** You now have 17 gallons in the 17-gallon pail and 6 gallons in the 9-gallon pail.

18. **Top off the 9-gallon pail using the 17-gallon pail.** Since the 9-gallon pail already has 6 gallons, it only needs 3 more gallons to be full. Pour water from the 17-gallon pail into the 9-gallon pail until it is full. This leaves you with 17 - 3 = 14 gallons of water in the 17-gallon pail.

19. **Empty the 9-gallon pail back into the river.** Now, you have 14 gallons in the 17-gallon pail.

20. **Fill the 9-gallon pail from the 17-gallon pail.** This leaves you with 14 - 9 = 5 gallons in the 17-gallon pail.

Now, you have exactly 5 gallons of water in the 17-gallon pail, which was the goal.
\end{lstlisting}

\subsection*{One-shot, Chain-of-Thought Problem-Space Formulation: $F(9,17)\rightarrow5$}
This formulation is correct but marginally precise enough to be used for execution. The transfer operator's post-conditions are described in natural language rather than formal annotation and would not be sufficient to generate an operator implementation for an executable agent.

\begin{lstlisting}[breaklines]
### Problem Space Formulation for the General Problem

#### States
- A state can be represented as a tuple \((x, y)\), where \(x\) and \(y\) are the current amounts of water in the two containers, respectively.

#### Initial State
- The initial state is \((0, 0)\), assuming both containers start empty.

#### Goal State
- The goal state is defined as any state \((a, b)\) where \(a + b = G\), and \(G\) is the specific amount of water needed.

#### Operators
1. **Fill**: Filling a container to its full capacity.
   - **Precondition**: Container is not full.
   - **Postcondition**: Container is full, i.e., if the first container's capacity is \(C_1\), then after filling, the state is \((C_1, y)\) or \((x, C_2)\) for the second container with capacity \(C_2\).
   
2. **Empty**: Emptying a container completely.
   - **Precondition**: Container is not empty.
   - **Postcondition**: Container is empty, resulting in a state \((0, y)\) or \((x, 0)\).
   
3. **Transfer**: Transferring water from one container to the other until the first is empty or the second is full.
   - **Precondition**: The source container is not empty, and the destination container is not full.
   - **Postcondition**: Water is transferred from one container to the other, resulting in a new state \((x', y')\), where \(x'\) and \(y'\) depend on the initial amounts and the capacities of the containers.

#### Illegal or Undesirable States
- Any state where the amount of water in either container exceeds its capacity or is negative is considered illegal.
- States where the total amount of water does not move towards achieving the goal (e.g., unnecessary emptying and refilling cycles that do not change the total amount of water held) can be considered undesirable.

This formulation captures the essence of the problem space for solving water measurement problems with two containers of arbitrary capacities and a specific target volume.
\end{lstlisting}

\subsection*{CTA Agent: $F(9,17)\rightarrow5$}
This formulation output is correct and sufficiently precise to support execution. The search control output, however, expresses the idea that there should be monotonic progress toward the goal, which is incorrect for water jugs generally (and this problem specifically). The concept of monotonic progress toward the goal comes up in several other CoT outcomes, but generally has not been common in the CTA Agent outcomes. One possible theory might be that the larger volumes (and the much less frequent appearance in training data, if this combination appears at all), causes the LLM to revert to more general patterns/rules of thumb.

\subsubsection*{Input Prompt}
\begin{lstlisting}[breaklines]

You are an expert in cognitive task analysis. You are helping to design a reasoner/problem solver that can solve many different instances of a class of problem. Problems are defined in 1-2 paragraphs. There will be a specific example problem provided. However,  your responses should focus creating a general formulation of the problem space, not a problem space specific to the example only. Focus each response on the most recent, specific question asked of you. The questions are designed to break the problem-space formulation into a set of discrete steps. Unless otherwise directed, be concise in each response (i.e., respond in one sentence or expression).

Use these definitions:
Problem Space: A problem space consists a set of symbolic structures (the states of the space) and a set of operators over the space. Each operator takes a state as input and produces a state as output (although there may be other inputs and outputs as well). The operators may be partial (i.e., not defined for all states). Sequences of operators define paths that thread their way through sequences of states. 
         
Problem: A problem in a problem space consists of a set of initial states, a set of goal states, and a set of path constraints. The problem is to find a path through the space that starts at the initial state, passes only along paths that satisfy the path constraints, and ends at any goal state. 

PROBLEM DESCRIPTION:

The general problem is to deliver a specific amount of water using containers that are opaque and have contain no graduated markings. The amount of water that needs to be delivered will generally differ from the full capacities of the container. As a specific example, how can you bring up from the river exactly 5 gallons of water when you have only two containers, a 17-gallon pail and a 9-gallon pail to measure with?

The solution to this problem will be solved by a search from the initial state to the goal state using the operators specified thus far.Search can be made more efficient by identfying unproductive paths such as loops and dead-ends. Your task is to identify/specify these. 
What are undesirable states? For example, generating a state that is identical to the initial state is often not a productive step.
What are undesirable sequences of operator applications? For example, an operator sequence that undoes the action of the immediately previous operator is often not desirable.Use formal notation and/or mathematical expressions to characterize any undesirable states or operator sequences.


LIMIT YOUR RESPONSE TO JUST THIS QUESTION/ASPECT OF ANALYSIS.
\end{lstlisting}

\subsubsection*{Initial Problem Characterization}
\begin{lstlisting}[breaklines]
Initial State: Two containers, a 17-gallon pail and a 9-gallon pail, both empty (0 gallons in each).

Final State: Exactly 5 gallons of water in one of the containers.

Illegal/Impossible States: Any state where the amount of water in either container exceeds its capacity (i.e., >17 gallons in the 17-gallon pail or >9 gallons in the 9-gallon pail) or is negative.
\end{lstlisting}

\subsubsection*{Refinement of Problem Characterization}
\begin{lstlisting}[breaklines]
The revised problem characterization, focusing on precision and clarity, is as follows:

Initial State: \(S_{initial} = \{(x=0, y=0)\}\), where \(x\) represents the volume of water in the 17-gallon pail and \(y\) represents the volume of water in the 9-gallon pail, both measured in gallons.

Goal State: \(S_{goal} = \{(x=5, y=0) \lor (x=0, y=5)\}\), indicating that the goal is achieved when either the 17-gallon pail (\(x\)) or the 9-gallon pail (\(y\)) contains exactly 5 gallons of water, with the other container's volume being irrelevant to the goal condition.

Illegal States: Defined by the conditions \(x > 17\), \(x < 0\), \(y > 9\), or \(y < 0\), representing states where the volume of water in either container exceeds its capacity or is negative, which are physically impossible.
\end{lstlisting}

\subsubsection*{Initial Operator/Problem-Space Characterization}
\begin{lstlisting}[breaklines]
Operators and their conditions in the problem space are as follows:

1. **Fill** an empty container to its full capacity:
   - **Precondition**: Container \(C\) is not full, \(C \in \{9, 17\}\) gallons.
   - **Effect**: Container \(C\) becomes full, \(State(C) = Capacity(C)\).

2. **Empty** a container:
   - **Precondition**: Container \(C\) is not empty.
   - **Effect**: Container \(C\) becomes empty, \(State(C) = 0\).

3. **Transfer** water from one container to another until the source is empty or the destination is full:
   - **Precondition**: Container \(A\) is not empty, and Container \(B\) is not full, \(A, B \in \{9, 17\}\), \(A \neq B\).
   - **Effect**: Water is transferred from \(A\) to \(B\) until \(A\) is empty or \(B\) is full, whichever comes first. The new state is \(State(A) = max(0, State(A) - (Capacity(B) - State(B)))\), \(State(B) = min(Capacity(B), State(B) + State(A))\).

These operators allow for the manipulation of water volumes within the constraints of the containers' capacities and the goal of measuring out a specific amount of water.
\end{lstlisting}

\subsubsection*{Refinement of Operator/Problem-Space Characterization}
\begin{lstlisting}[breaklines]
Upon review, the operators need refinement for clarity and completeness, particularly in specifying the preconditions and postconditions accurately. Here is the revised problem space characterization:

1. **Fill** a container to its full capacity:
   - **Precondition**: \(State(C) < Capacity(C)\), where \(C\) is either the 17-gallon or the 9-gallon container.
   - **Postcondition**: \(State(C) = Capacity(C)\).

2. **Empty** a container:
   - **Precondition**: \(State(C) > 0\).
   - **Postcondition**: \(State(C) = 0\).

3. **Transfer** water from one container to another:
   - **Precondition**: \(State(A) > 0\) and \(State(B) < Capacity(B)\), where \(A\) and \(B\) are distinct containers.
   - **Postcondition**: If \(State(A) + State(B) \leq Capacity(B)\), then \(State(B)' = State(A) + State(B)\) and \(State(A)' = 0\). If \(State(A) + State(B) > Capacity(B)\), then \(State(A)' = State(A) - (Capacity(B) - State(B))\) and \(State(B)' = Capacity(B)\).

These revisions ensure that the operators are defined for all relevant states in the problem space, accurately reflecting the possible actions (fill, empty, transfer) and their effects on the states of the containers.
\end{lstlisting}

\subsubsection*{Generation of Search Control}
\begin{lstlisting}[breaklines]
Undesirable states and sequences in this problem space include:

1. **Undesirable States**:
   - States where the total volume of water in both containers does not bring us closer to the goal state, specifically when the sum of water in both containers is less than 5 gallons after an operation, or exceeds the sum of the goal state and one container's capacity (e.g., \(x + y > 22\) for a goal of 5 gallons, considering a 17-gallon and a 9-gallon container).

2. **Undesirable Sequences of Operator Applications**:
   - Repeatedly filling and then emptying the same container without transferring water to the other container (\(Fill(C) \rightarrow Empty(C) \rightarrow Fill(C)\) without a \(Transfer\) in between).
   - Transferring water back and forth between containers without changing the state towards the goal (\(Transfer(A, B) \rightarrow Transfer(B, A)\) without any other operation that moves towards the goal state).
   - Any sequence that leads to a state previously encountered, as this indicates a loop.

Formally, we can denote undesirable sequences as those that include operations \(O_i\) and \(O_{i+1}\) where \(O_i\) undoes the effect of \(O_{i+1}\) or vice versa, specifically:
- If \(O_i = Fill(C)\) and \(O_{i+1} = Empty(C)\), or \(O_i = Empty(C)\) and \(O_{i+1} = Fill(C)\), without a \(Transfer\) operation to the other container in between, this sequence is undesirable.
- If \(O_i = Transfer(A, B)\) and \(O_{i+1} = Transfer(B, A)\) resulting in a state \(S_j\) that is identical to a previous state \(S_k\), where \(j > k\), this indicates a loop and is undesirable.

These guidelines help in pruning the search space by avoiding actions that do not contribute to reaching the goal state more efficiently.
\end{lstlisting}

\clearpage
\section*{Variation: $V(4q_{uart},9g_{allon})\rightarrow6g_{allon}$}

\subsection*{$V(4q,9g)\rightarrow6g$ Problem Description}

The general problem is to deliver a specific amount of water using containers that are opaque and have contain no graduated markings. \
The amount of water that needs to be delivered will generally differ from the full capacities of the container. As a specific example, \
how can you bring up from the river exactly 6 gallons of water when you have only two containers, \
a 4-quart pail and a 9-gallon pail to measure with?

\subsection*{Soar Agent Solution}

\begin{lstlisting}[breaklines]
     1:    O: O1 (initialize-water-jug-look-ahead)
S1: 1: 0
S1: 9: 0
     2:    O: O3 (fill)
S1:  FILL(9)
S1: 9: 9
     3:    O: O4 (pour)
S1:  POUR(9:9,1:0)
S1: 1: 1
S1: 9: 8
     4:    O: O6 (empty)
S1:  EMPTY(1)
S1: 1: 0
     5:    O: O12 (pour)
S1:  POUR(9:8,1:0)
S1: 1: 1
S1: 9: 7
     6:    O: O14 (empty)
S1:  EMPTY(1)
S1: 1: 0
     7:    O: O20 (pour)
S1:  POUR(9:7,1:0)
S1: 1: 1
S1: 9: 6
S1 Solution Found!

 
\end{lstlisting}

\subsection*{One-shot, Chain-of-Thought Solution: $V(4q,9g)\rightarrow6g$}
This result is somewhat interesting in its incorrectness. It immediately identifies the 4-quart pail as identical to a 1-gallon pail and correctly chooses to start by filling the 9-gallon pail (the shortest solution to this problem is to fill the 9-gallon pail and then transfer 1-gallon to the other pail three times, emptying the 1-gallon pail after each pour). The mistake comes in step 4, where the LLM does not recognize that the 6-gallon state is reached.

\begin{lstlisting}[breaklines]

To solve this problem, we need to use the two containers we have a 4-quart pail and a 9-gallon pail to measure out exactly 6 gallons of water. It is important to note that 1 gallon equals 4 quarts, so the 4-quart pail is essentially a 1-gallon pail. Here's a step-by-step solution:

1. **Fill the 9-gallon pail to the top with water from the river.** This is your starting point.

2. **Use the 9-gallon pail to fill the 4-quart (1-gallon) pail.** After doing this, you will have 8 gallons of water left in the 9-gallon pail.

3. **Empty the 4-quart (1-gallon) pail back into the river.** Now, you're ready to measure more water.

4. **Repeat steps 2 and 3 three more times.** Each time, you transfer 1 gallon from the 9-gallon pail to the 4-quart pail and then empty it, reducing the water in the 9-gallon pail by 1 gallon each time. After doing this three more times, you will have transferred and emptied a total of 4 gallons, leaving you with 4 gallons in the 9-gallon pail.

5. **Fill the 4-quart (1-gallon) pail again from the 9-gallon pail.** Now, you have 3 gallons left in the 9-gallon pail.

6. **Empty the 4-quart (1-gallon) pail back into the river, and then fill it again from the 9-gallon pail.** This leaves you with exactly 2 gallons in the 9-gallon pail.

7. **Pour the 2 gallons from the 9-gallon pail into a separate container (if available) or carefully hold it if another container is not available.** 

8. **Fill the 9-gallon pail to the top again and add the 2 gallons you set aside to the 9-gallon pail.** Now, you have exactly 11 gallons in the 9-gallon pail.

9. **Use the 9-gallon pail to fill the 4-quart (1-gallon) pail five times, emptying it each time back into the river.** This process will remove 5 gallons from the 9-gallon pail, leaving you with exactly 6 gallons.

This solution effectively uses the tools available to measure out exactly 6 gallons of water, despite the containers not being initially suited for the exact measurement needed.


\end{lstlisting}

\subsection*{One-shot, Chain-of-Thought Problem-Space Formulation: $V(4q,9g)\rightarrow6g$}

This solution is underspecified and insufficient to form a correct, executable set of operators for search. The units of the pails (which are different from one another in this variation) are never referenced in the problem formulation at all. Similar to other results from one-shot, CoT problem formulation, the transfer operator is incompletely specified which is an critical gap for the mixed units version of the problem in particular.

\begin{lstlisting}[breaklines]

### Problem Space Formulation for the General Problem

#### States
- A state can be represented as a tuple \((V_1, V_2)\), where \(V_1\) and \(V_2\) are the volumes of water in the two containers at any point in time.

#### Initial State
- The initial state is \((0, 0)\), assuming both containers start empty.

#### Goal State
- The goal state is defined as any state \((V_1, V_2)\) where \(V_1 + V_2 = G\) and \(G\) is the specific amount of water needed.

#### Operators
1. **Fill**: Fills a container to its full capacity.
   - **Preconditions**: Container is not full.
   - **Postconditions**: Container is full.
2. **Empty**: Empties a container completely.
   - **Preconditions**: Container is not empty.
   - **Postconditions**: Container is empty.
3. **Transfer**: Transfers water from one container to the other until the source container is empty or the destination container is full.
   - **Preconditions**: The source container is not empty, and the destination container is not full.
   - **Postconditions**: Water volume is transferred from one container to the other.

#### Illegal or Undesirable States
- Any state where \(V_1\) or \(V_2\) exceeds the capacity of its respective container is illegal.
- Any state where \(V_1\) or \(V_2\) is negative is illegal.

This formulation captures the essence of the problem space for solving water delivery problems with containers of unmarked capacities, focusing on the operations that can be performed and the conditions that govern these operations.
\end{lstlisting}

\subsection*{CTA Agent: $V(4q,9g)\rightarrow6g$}

The CTA agent successfully formulates the problem space, recognizing and adapting to the differences in the units of the jugs. More specifically, it defines the state value for the 4-quart jug as ``the volume in gallons in the 4-quart pail... Note: 4 quarts = 1 gallon.'' From that point forward, it produces descriptions of operators that are consistent that that definition. The search control is generally correct except for the notion that monotonic progress toward 6 gallons is expected.

\subsubsection*{Input Prompt}
\begin{lstlisting}[breaklines]
You are an expert in cognitive task analysis. You are helping to design a reasoner/problem solver that can solve many different instances of a class of problem. Problems are defined in 1-2 paragraphs. There will be a specific example problem provided. However,  your responses should focus creating a general formulation of the problem space, not a problem space specific to the example only. Focus each response on the most recent, specific question asked of you. The questions are designed to break the problem-space formulation into a set of discrete steps. Unless otherwise directed, be concise in each response (i.e., respond in one sentence or expression).

Use these definitions:
Problem Space: A problem space consists a set of symbolic structures (the states of the space) and a set of operators over the space. Each operator takes a state as input and produces a state as output (although there may be other inputs and outputs as well). The operators may be partial (i.e., not defined for all states). Sequences of operators define paths that thread their way through sequences of states. 
         
Problem: A problem in a problem space consists of a set of initial states, a set of goal states, and a set of path constraints. The problem is to find a path through the space that starts at the initial state, passes only along paths that satisfy the path constraints, and ends at any goal state. 

PROBLEM DESCRIPTION:

The general problem is to deliver a specific amount of water using containers that are opaque and have contain no graduated markings. The amount of water that needs to be delivered will generally differ from the full capacities of the container. As a specific example, how can you bring up from the river exactly 6 gallons of water when you have only two containers, a 4-quart pail and a 9-gallon pail to measure with?

The solution to this problem will be solved by a search from the initial state to the goal state using the operators specified thus far.Search can be made more efficient by identfying unproductive paths such as loops and dead-ends. Your task is to identify/specify these. 
What are undesirable states? For example, generating a state that is identical to the initial state is often not a productive step.
What are undesirable sequences of operator applications? For example, an operator sequence that undoes the action of the immediately previous operator is often not desirable.Use formal notation and/or mathematical expressions to characterize any undesirable states or operator sequences.

LIMIT YOUR RESPONSE TO JUST THIS QUESTION/ASPECT OF ANALYSIS.
\end{lstlisting}

\subsubsection*{Initial Problem Characterization}
\begin{lstlisting}[breaklines]
Initial State: Both containers are empty, represented as (0, 0), where the first value in the tuple represents the amount of water in the 4-quart pail (with a maximum capacity of 1 gallon, since 4 quarts = 1 gallon), and the second value represents the amount of water in the 9-gallon pail.

Final State: The goal is to have exactly 6 gallons of water in one of the containers, which can be represented as either (x, 6) or (6, y), where x and y can be any value that does not exceed the respective container's capacity.

Illegal/Impossible States: Any state where the amount of water in either container exceeds its capacity is illegal. Thus, states where the first value is greater than 1 or the second value is greater than 9 are not allowed. Additionally, since the problem specifies delivering a specific amount of water, any state that does not contribute to achieving exactly 6 gallons in one of the containers could be considered irrelevant or outside the problem space, depending on the path constraints defined later.
\end{lstlisting}

\subsubsection*{Refinement of Problem Characterization}
\begin{lstlisting}[breaklines]
The revised problem characterization, focusing on precision and clarity, is as follows:

Initial State: \(S_{initial} = (0, 0)\), where the first element of the tuple represents the volume in gallons in the 4-quart pail (\(P_{4qt}\)) and the second element represents the volume in gallons in the 9-gallon pail (\(P_{9gal}\)). Note: 4 quarts = 1 gallon.

Goal State: \(S_{goal} = \{(x, 6), (6, y)\}\) for \(0 \leq x \leq 1\) and \(0 \leq y \leq 9\), indicating that the goal is achieved when either container holds exactly 6 gallons of water, regardless of the amount in the other container.

Constraints on States: Given the capacities, the set of legal states \(S\) is defined by \(0 \leq x \leq 1\) and \(0 \leq y \leq 9\), where \(x\) and \(y\) represent the volume in gallons in \(P_{4qt}\) and \(P_{9gal}\), respectively. Any state outside these bounds is considered illegal.

This characterization ensures clarity by specifying the capacities and constraints in quantitative terms, focusing on the exact volumes that define the initial, goal, and legal states within the problem space.

\end{lstlisting}

\subsubsection*{Initial Operator/Problem-Space Characterization}
\begin{lstlisting}[breaklines]
The operators relevant to solving this problem, along with their preconditions and postconditions, are as follows:

1. **Fill Operator** (\(Fill_i\)): Fills container \(i\) to its maximum capacity.
   - **Precondition**: Container \(i\) is not full.
   - **Postcondition**: Container \(i\) is full, represented as \(i = Capacity_i\).

2. **Empty Operator** (\(Empty_i\)): Empties container \(i\).
   - **Precondition**: Container \(i\) contains some water (\(i > 0\)).
   - **Postcondition**: Container \(i\) is empty, represented as \(i = 0\).

3. **Transfer Operator** (\(Transfer_{i \rightarrow j}\)): Transfers water from container \(i\) to container \(j\) until container \(i\) is empty or container \(j\) is full, whichever comes first.
   - **Precondition**: Container \(i\) is not empty (\(i > 0\)) and container \(j\) is not full (\(j < Capacity_j\)).
   - **Postcondition**: The new state of container \(i\) and \(j\) is updated based on the amount of water transferred, represented as \(i' = max(0, i - (Capacity_j - j))\) and \(j' = min(Capacity_j, j + i)\).

These operators define the actions that can be taken at any point in the problem space, with the preconditions specifying when an operator can be applied and the postconditions describing the state of the system after the operator is applied.
\end{lstlisting}

\subsubsection*{Refinement of Operator/Problem-Space Characterization}
\begin{lstlisting}[breaklines]
Upon reviewing the initial operator definitions, adjustments are necessary to ensure completeness and accuracy in the context of the problem space. Here is the revised set of operators with corrected preconditions and postconditions:

1. **Fill Operator** (\(Fill_i\)): Fills container \(i\) to its maximum capacity from an unlimited water source.
   - **Precondition**: \(i < Capacity_i\), where \(i\) is the current volume in container \(i\) and \(Capacity_i\) is its maximum capacity.
   - **Postcondition**: \(i = Capacity_i\).

2. **Empty Operator** (\(Empty_i\)): Empties container \(i\) completely.
   - **Precondition**: \(i > 0\), indicating that there is some volume of water in container \(i\).
   - **Postcondition**: \(i = 0\).

3. **Transfer Operator** (\(Transfer_{i \rightarrow j}\)): Transfers water from container \(i\) to container \(j\).
   - **Precondition**: \(i > 0\) and \(j < Capacity_j\), indicating that container \(i\) has water to transfer and container \(j\) has room to receive water.
   - **Postcondition**: \(i' = max(0, i - (Capacity_j - j))\) and \(j' = min(Capacity_j, j + i)\), ensuring that water is transferred from \(i\) to \(j\) until \(i\) is empty or \(j\) is full, whichever occurs first.

These revised operators more accurately reflect the actions available within the problem space, ensuring that the preconditions and postconditions align with the logical constraints of the problem.
\end{lstlisting}

\subsubsection*{Generation of Search Control}
\begin{lstlisting}[breaklines]
Undesirable states and sequences in this problem space include:

1. **Undesirable States**: Any state where the volumes in both containers do not progress towards achieving the goal state of having exactly 6 gallons in one of the containers. Specifically, states where the total volume of water does not change after an operation, or where the operation results in a volume that has previously been discarded as unhelpful, can be considered undesirable. For example, states like \((1, 9)\) or \((0, 9)\) after attempting to solve for 6 gallons are not directly useful.

2. **Undesirable Sequences of Operator Applications**:
   - **Redundant Fill**: Applying a Fill operator to a container that is already full, i.e., \(Fill_i\) followed by another \(Fill_i\) without any intermediate operation that changes the state of \(i\).
   - **Redundant Empty**: Applying an Empty operator to a container that is already empty, i.e., \(Empty_i\) followed by another \(Empty_i\) without any intermediate operation that changes the state of \(i\).
   - **Canceling Operations**: A sequence where a Fill or Transfer operation is immediately followed by an Empty operation on the same container, effectively undoing the previous action without making progress towards the goal, e.g., \(Fill_i; Empty_i\) or \(Transfer_{i \rightarrow j}; Empty_j\).
   - **Cyclical Transfers**: A sequence of Transfer operations that return the system to a previous state without achieving the goal, e.g., \(Transfer_{i \rightarrow j}; Transfer_{j \rightarrow i}\) repeated without any progress.

Formally, these can be characterized as follows:
- For any state \(S = (x, y)\), an undesirable state is one where \(x + y\) does not incrementally approach 6 gallons without exceeding the capacities of either container.
- For sequences, if \(O_n\) and \(O_{n+1}\) are operations such that the state post-\(O_{n+1}\) is equivalent to the state pre-\(O_n\), then the sequence \([O_n, O_{n+1}]\) is undesirable.
\end{lstlisting}

\clearpage
\section*{Variation: $V(2,3,5)\rightarrow4$}

\subsection*{$V(2,3,5)\rightarrow4$ Problem Description}

The general problem is to deliver a specific amount of water using containers that are opaque and have contain no graduated markings. \
The amount of water that needs to be delivered will generally differ from the full capacities of the container. As a specific example, \
how can you bring up from the river exactly 4 gallons of water when you have only three containers, \
a 5-gallon pail, a 3-gallon pail, and a 2-gallon pail to measure with?

\subsection*{Soar Agent Solution}

\begin{lstlisting}[breaklines]
     1:    O: O1 (initialize-water-jug-look-ahead)
S1: 2: 0
S1: 3: 0
S1: 5: 0
     2:    O: O2 (fill)
S1:  FILL(2)
S1: 2: 2
     3:    O: O5 (pour)
S1:  POUR(2:2,5:0)
S1: 5: 2
S1: 2: 0
     4:    O: O11 (fill)
S1:  FILL(2)
S1: 2: 2
     5:    O: O14 (pour)
S1:  POUR(2:2,5:2)
S1: 5: 4
S1 Solution Found!
S1: 2: 0
 
\end{lstlisting}

\subsection*{One-shot, Chain-of-Thought Solution: $V(2,3,5)\rightarrow4$}

This solution is correct, but follows the familiar pattern and ignores the second jug. The resulting plan takes 50\% more steps than the one that uses the 2-gallon jug.

\begin{lstlisting}[breaklines]

To solve this problem, we can use a method called the "Water Jug Problem" or "Die Hard Water Puzzle" which involves filling and transferring water between the containers to achieve the desired amount. Here's a step-by-step solution:

1. Fill the 5-gallon pail to the top.
2. Pour water from the 5-gallon pail into the 3-gallon pail until the 3-gallon pail is full. This leaves 2 gallons of water in the 5-gallon pail.
3. Empty the 3-gallon pail.
4. Pour the remaining 2 gallons of water from the 5-gallon pail into the empty 3-gallon pail.
5. Fill the 5-gallon pail to the top again.
6. Pour water from the 5-gallon pail into the 3-gallon pail until the 3-gallon pail is full. This leaves 4 gallons of water in the 5-gallon pail.

By following these steps, you can measure exactly 4 gallons of water using the 5-gallon, 3-gallon, and 2-gallon pails. This method utilizes the capacities of the containers to manipulate the water levels effectively.
\end{lstlisting}

\subsection*{One-shot, Chain-of-Thought Problem-Space Formulation: $V(2,3,5)\rightarrow4$}

This problem-space formulation is incomplete but otherwise correct. Incompleteness includes the lack of specificity in the specification of the transfer operator. Further, only illegal states are indicated rather than also indicating non-desirable search states.

\begin{lstlisting}[breaklines]
Initial state: (0, 0, 0) - all pails are empty.
Goal state: (x, y, z) where x + y + z = 4 (the total amount of water needed).

Operators:
1. Fill(pail): Fills the specified pail to its maximum capacity.
   Preconditions: Pail is not already full.
   Postconditions: Pail is full.

2. Empty(pail): Empties the specified pail.
   Preconditions: Pail is not already empty.
   Postconditions: Pail is empty.

3. Transfer(from, to): Transfers water from one pail to another until either the "from" pail is empty or the "to" pail is full.
   Preconditions: "from" pail is not empty, "to" pail is not full.
   Postconditions: Water is transferred from "from" pail to "to" pail.

Illegal states: Any state where the amount of water in a pail exceeds its capacity or is negative.
\end{lstlisting}

\subsection*{CTA Agent: $V(2,3,5)\rightarrow4$}
This formulation is correct and precise and accurately (only after refinement) reflects the addition of the third pail.

\subsubsection*{Input Prompt}
\begin{lstlisting}[breaklines]
You are an expert in cognitive task analysis. You are helping to design a reasoner/problem solver that can solve many different instances of a class of problem. Problems are defined in 1-2 paragraphs. There will be a specific example problem provided. However,  your responses should focus creating a general formulation of the problem space, not a problem space specific to the example only. Focus each response on the most recent, specific question asked of you. The questions are designed to break the problem-space formulation into a set of discrete steps. Unless otherwise directed, be concise in each response (i.e., respond in one sentence or expression).

Use these definitions:
Problem Space: A problem space consists a set of symbolic structures (the states of the space) and a set of operators over the space. Each operator takes a state as input and produces a state as output (although there may be other inputs and outputs as well). The operators may be partial (i.e., not defined for all states). Sequences of operators define paths that thread their way through sequences of states. 
         
Problem: A problem in a problem space consists of a set of initial states, a set of goal states, and a set of path constraints. The problem is to find a path through the space that starts at the initial state, passes only along paths that satisfy the path constraints, and ends at any goal state. 

PROBLEM DESCRIPTION:

The general problem is to deliver a specific amount of water using containers that are opaque and have contain no graduated markings. The amount of water that needs to be delivered will generally differ from the full capacities of the container. As a specific example, how can you bring up from the river exactly 4 gallons of water when you have only three containers, a 5-gallon pail, a 3-gallon pail, and a 2-gallon pail to measure with?

The solution to this problem will be solved by a search from the initial state to the goal state using the operators specified thus far.Search can be made more efficient by identfying unproductive paths such as loops and dead-ends. Your task is to identify/specify these. 
What are undesirable states? For example, generating a state that is identical to the initial state is often not a productive step.
What are undesirable sequences of operator applications? For example, an operator sequence that undoes the action of the immediately previous operator is often not desirable.Use formal notation and/or mathematical expressions to characterize any undesirable states or operator sequences.

LIMIT YOUR RESPONSE TO JUST THIS QUESTION/ASPECT OF ANALYSIS.
\end{lstlisting}

\subsubsection*{Initial Problem Characterization}
\begin{lstlisting}[breaklines]
Initial State: (0, 0, 0) representing the amount of water in the 5-gallon, 3-gallon, and 2-gallon pails respectively.

Final State: (x, y, 4) where x and y can be any value such that the total amount of water in all containers equals 4 gallons. The primary goal is to have a total of 4 gallons, not necessarily in one container.

Illegal/Impossible States: Any state where the amount of water in any of the containers exceeds its capacity or is negative, i.e., states (x, y, z) where x > 5, y > 3, z > 2, or any of x, y, z < 0 are not allowed.
\end{lstlisting}

\subsubsection*{Refinement of Problem Characterization}
\begin{lstlisting}[breaklines]
The revised problem characterization, focusing on precision and completeness, is as follows:

Initial State: \(S_{initial} = (0, 0, 0)\), where each value represents the volume of water in the 5-gallon, 3-gallon, and 2-gallon containers respectively.

Goal State(s): \(S_{goal} = \{(x, y, z) | x + y + z = 4\}\), where \(x \leq 5\), \(y \leq 3\), and \(z \leq 2\). This set of states reflects all combinations where the total volume of water equals exactly 4 gallons, irrespective of the distribution across the containers.

Constraints on States: For any state \(S = (x, y, z)\),
- \(0 \leq x \leq 5\), indicating the volume in the 5-gallon container cannot exceed its capacity or be negative.
- \(0 \leq y \leq 3\), indicating the volume in the 3-gallon container cannot exceed its capacity or be negative.
- \(0 \leq z \leq 2\), indicating the volume in the 2-gallon container cannot exceed its capacity or be negative.

This characterization removes ambiguity by specifying the capacities and constraints explicitly and defines the goal state in terms of the total volume of water across all containers, ensuring clarity in the problem's objectives.
\end{lstlisting}

\subsubsection*{Initial Operator/Problem-Space Characterization}
\begin{lstlisting}[breaklines]
Operators for the problem space, along with their preconditions and postconditions, are as follows:

1. **Fill** a container from the river:
   - **Operator**: \(Fill(i)\)
   - **Preconditions**: Container \(i\) is not full.
   - **Postconditions**: Container \(i\) is full.

2. **Empty** a container into the river:
   - **Operator**: \(Empty(i)\)
   - **Preconditions**: Container \(i\) contains some water.
   - **Postconditions**: Container \(i\) is empty.

3. **Transfer** water from one container to another:
   - **Operator**: \(Transfer(i, j)\)
   - **Preconditions**: Container \(i\) contains some water, and container \(j\) is not full.
   - **Postconditions**: Water is transferred from container \(i\) to container \(j\) until either container \(i\) is empty or container \(j\) is full, whichever comes first.

In these operators, \(i\) and \(j\) are identifiers for the containers (e.g., 1 for the 5-gallon, 2 for the 3-gallon, and 3 for the 2-gallon containers). These operators are defined for all states where their preconditions are met, making them partial functions of the problem space.
\end{lstlisting}

\subsubsection*{Refinement of Operator/Problem-Space Characterization}
\begin{lstlisting}[breaklines]
Upon reviewing the initial operator definitions, adjustments and clarifications are necessary to ensure completeness and correctness:

1. **Fill** a container from the river:
   - **Operator**: \(Fill(i)\)
   - **Preconditions**: \(Volume(i) < Capacity(i)\), where \(Volume(i)\) is the current volume of water in container \(i\), and \(Capacity(i)\) is the maximum capacity of container \(i\).
   - **Postconditions**: \(Volume(i) = Capacity(i)\), setting the volume of container \(i\) to its maximum capacity.

2. **Empty** a container into the river:
   - **Operator**: \(Empty(i)\)
   - **Preconditions**: \(Volume(i) > 0\), indicating that container \(i\) contains some water.
   - **Postconditions**: \(Volume(i) = 0\), making container \(i\) empty.

3. **Transfer** water from one container to another:
   - **Operator**: \(Transfer(i, j)\)
   - **Preconditions**: \(Volume(i) > 0\) and \(Volume(j) < Capacity(j)\), indicating that container \(i\) has water to transfer and container \(j\) has space to receive water.
   - **Postconditions**: \(Volume(i) = max(0, Volume(i) - (Capacity(j) - Volume(j)))\) and \(Volume(j) = min(Capacity(j), Volume(j) + Volume(i))\), reflecting the new volumes of containers \(i\) and \(j\) after the transfer. This ensures that water is transferred from \(i\) to \(j\) until \(i\) is empty or \(j\) is full, whichever occurs first.

These revised operator definitions provide a clearer and more precise framework for the problem space, ensuring that the preconditions and postconditions accurately reflect the possible actions and their effects within the problem domain.
\end{lstlisting}

\subsubsection*{Generation of Search Control}
\begin{lstlisting}[breaklines]
Undesirable states and sequences in this problem space include:

1. **Undesirable States**:
   - States where the total volume of water across all containers does not move towards achieving the goal state, such as any state where the sum of volumes in all containers decreases or remains the same after an operation that does not contribute to reaching the goal state of having exactly 4 gallons in total.

2. **Undesirable Sequences of Operator Applications**:
   - Sequences that involve filling a container and then immediately emptying the same container without any intermediate beneficial operation, formally: \(Fill(i); Empty(i)\) or \(Empty(i); Fill(i)\), where \(i\) represents any container.
   - Sequences that involve transferring water from one container to another and then immediately transferring it back without any intermediate operation that changes the state beneficially, formally: \(Transfer(i, j); Transfer(j, i)\), where \(i\) and \(j\) represent different containers.
   - Any sequence of operations that returns to a previously visited state, creating a loop, can be considered unproductive. This can be formally represented as a sequence of operations \(O_1, O_2, ..., O_n\) that transforms the initial state \(S_0\) back to \(S_0\) or any intermediate state \(S_k\) (where \(0 < k < n\)) without reaching the goal state.

Identifying and avoiding these undesirable states and sequences can significantly improve the efficiency of the search process by reducing the number of explored states and focusing on those that are more likely to lead towards the goal state.
\end{lstlisting}

\clearpage
\section*{Analogue: $A(4,9)\rightarrow6$}

\subsection*{$A(4,9)\rightarrow6$ Problem Description}

You are an engineer on an alien spaceship. Physics works as in our universe but the technology differs significantly. \
The general problem is to deliver a specific amount of 
"flucotone" (a type of energy) using flucotone-holding devices (FHDs). 
Flucotone can be transferred to FHDs up to their specified capacity with no energy/flucotone loss. \
An FHD can hold any amount of flucotone up its capacity but there is no way to measure how much flucotone is in an FHD other than it being full or empty. \
The amount of flucotone needed for any specific task will generally differ from the full capacities of available FHDs. As a specific example, \
assume a power plant that can generate limitless flucotone. From this power plant, how can you deliver exactly 6 units of flucotone to the engine room \ 
when you have only two FHDs: a 4-unit FHD and a 9-unit FHD?

\subsection*{Soar Agent Solution}
The Soar solution for this problem is identical to the $F(4,9)\rightarrow6$ solution.

\subsection*{One-shot, Chain-of-Thought Solution: $A(4,9)\rightarrow6$}

This solution is correct.

\begin{lstlisting}[breaklines]

To solve the problem of delivering exactly 6 units of flucotone using a 4-unit FHD and a 9-unit FHD, follow these steps:

1. **Fill the 9-unit FHD** with flucotone from the power plant.

2. **Transfer from the 9-unit FHD to the 4-unit FHD** until the 4-unit FHD is full. This leaves you with 5 units of flucotone in the 9-unit FHD (since you've transferred 4 units out).

3. **Empty the 4-unit FHD** back into the power plant (or another storage if that's the process).

4. **Transfer the remaining 5 units from the 9-unit FHD to the 4-unit FHD**. Now, the 4-unit FHD has 5 units, but since its capacity is only 4 units, it means you've actually only transferred 4 units, leaving 1 unit in the 9-unit FHD.

5. **Fill the 9-unit FHD again** from the power plant.

6. **Transfer from the 9-unit FHD to the 4-unit FHD** until the 4-unit FHD is full. Since the 4-unit FHD already has 1 unit, you only need to transfer 3 more units to fill it. This leaves you with 6 units of flucotone in the 9-unit FHD (9 - 3 = 6).

7. **Deliver the 6 units of flucotone in the 9-unit FHD** to the engine room.

By following these steps, you've successfully delivered exactly 6 units of flucotone to the engine room using only a 4-unit FHD and a 9-unit FHD.

\end{lstlisting}

\subsection*{One-shot, Chain-of-Thought Problem-Space Formulation: $A(4,9)\rightarrow6$}

This formulation is correct but incomplete/not sufficient for execution in a problem solver. In particular, the transfer (``pour" analogue) does not specify how to update the values in the containers. Additionally, the illegal/undesirable states are generic and would not reduce search generally (e.g., avoiding waste is counterproductive to solving a water-jugs problem).

\begin{lstlisting}[breaklines]

### Problem Space Formulation for the General Problem

#### States
A state can be represented as a tuple showing the amount of flucotone in each FHD and the amount needed to be delivered. For example, with two FHDs and a requirement to deliver a specific amount of flucotone, a state could be represented as \( (x, y, z) \), where:
- \( x \) is the amount of flucotone in the first FHD,
- \( y \) is the amount of flucotone in the second FHD,
- \( z \) is the amount of flucotone needed to be delivered.

#### Initial State
The initial state is when both FHDs are empty, and a specific amount of flucotone needs to be delivered. For example, \( (0, 0, Z) \), where \( Z \) is the total amount of flucotone needed.

#### Goal State
The goal state is any state where the total amount of flucotone delivered equals the amount required, regardless of the distribution in the FHDs. For example, \( (*, *, 0) \), where \( * \) can be any value representing the amount of flucotone in the FHDs, and the last element is 0, indicating the delivery requirement has been met.

#### Operators
Operators are actions that can change the state. For this problem, operators include:

1. **Fill**: Filling an FHD to its maximum capacity.
   - **Precondition**: FHD is not full.
   - **Postcondition**: FHD is full.
   
2. **Empty**: Emptying an FHD.
   - **Precondition**: FHD is not empty.
   - **Postcondition**: FHD is empty.
   
3. **Transfer**: Transferring flucotone from one FHD to another until the second is full or the first is empty.
   - **Precondition**: The source FHD is not empty, and the destination FHD is not full.
   - **Postcondition**: The amount of flucotone in the source and destination FHDs is adjusted according to the transfer.

#### Illegal or Undesirable States
- Any state where the amount of flucotone in an FHD exceeds its capacity.
- States where the total amount of flucotone exceeds the initial requirement plus the capacities of all FHDs, indicating wastage or inefficiency.

This formulation allows for the representation and manipulation of the problem space to find a sequence of operations that leads from the initial state to a goal state, satisfying the requirement of delivering a specific amount of flucotone using the available FHDs.
\end{lstlisting}

\subsection*{CTA Agent: $A(4,9)\rightarrow6$}
This formulation is correct and precise. The search control is not sufficiently formalized to support code generation but the content is generally correct.

\subsubsection*{Input Prompt}
\begin{lstlisting}[breaklines]
You are an expert in cognitive task analysis. You are helping to design a reasoner/problem solver that can solve many different instances of a class of problem. Problems are defined in 1-2 paragraphs. There will be a specific example problem provided. However,  your responses should focus creating a general formulation of the problem space, not a problem space specific to the example only. Focus each response on the most recent, specific question asked of you. The questions are designed to break the problem-space formulation into a set of discrete steps. Unless otherwise directed, be concise in each response (i.e., respond in one sentence or expression).

Use these definitions:
Problem Space: A problem space consists a set of symbolic structures (the states of the space) and a set of operators over the space. Each operator takes a state as input and produces a state as output (although there may be other inputs and outputs as well). The operators may be partial (i.e., not defined for all states). Sequences of operators define paths that thread their way through sequences of states. 
         
Problem: A problem in a problem space consists of a set of initial states, a set of goal states, and a set of path constraints. The problem is to find a path through the space that starts at the initial state, passes only along paths that satisfy the path constraints, and ends at any goal state. 

SPECIFIC PROBLEM:

You are an engineer on an alien spaceship. Physics works as in our universe but the technology differs significantly. The general problem is to deliver a specific amount of "flucotone" (a type of energy)) using flucotone-holding devices (FHDs). 
Flucotone can be transferred to FHDs up to their specified capacity with no energy/flucotone loss. An FHD can hold any amount of flucotone up its capacity but there is no way to measure how much flucontone is in an FHD other than it being full or empty. The amount of flucotone needed for any specific task will generally differ from the full capacities of available FHDs. As a specific example, assume a power plant that can generate limitless flucotone. From this power plant, how can you deliver exactly 6 units of flucotone to the engine room \ 
when you have only two FHDs: a 4-unit FHD and a 9-unit FHD?

For this response, focus on characterizing the problem itself. What is the initial state? What is the final state? Are there illegal / impossible states that are not allowed for this problem? If so, identify them.
Use formal notation and/or mathematical expressions if/as possible to characterize the problem. Be concise.

LIMIT YOUR RESPONSE TO THIS QUESTION ONLY/ASPECT OF ANALYSIS.
\end{lstlisting}

\subsubsection*{Initial Problem Characterization}
\begin{lstlisting}[breaklines]
Initial State: FHD1 = 0 units (4-unit capacity), FHD2 = 0 units (9-unit capacity), Goal = 6 units of flucotone to be delivered.

Final State: Any state where a total of 6 units of flucotone have been delivered to the engine room, irrespective of the distribution between FHD1 and FHD2.

Illegal/Impossible States: Any state where FHD1 > 4 units or FHD2 > 9 units, or the total flucotone delivered exceeds 6 units.
\end{lstlisting}

\subsubsection*{Refinement of Problem Characterization}
\begin{lstlisting}[breaklines]
The problem space can be more precisely characterized using the following formal notation:

- **Initial State (S0):** \( (FHD1 = 0, FHD2 = 0) \), where \(FHD1\) and \(FHD2\) represent the amount of flucotone in each Flucotone-Holding Device respectively.
- **Goal State (SG):** A state where the total amount of flucotone delivered equals 6 units. This can be represented as any state \( (FHD1, FHD2) \) where \( FHD1 + FHD2 = 6 \) and \( 0 \leq FHD1 \leq 4 \), \( 0 \leq FHD2 \leq 9 \).
- **Illegal States:** Any state where \( FHD1 > 4 \) or \( FHD2 > 9 \), or \( FHD1 + FHD2 \neq 6 \) when considering the goal state.

This characterization removes ambiguity by specifying the capacities and constraints of the FHDs and the exact goal condition in terms of flucotone units.
\end{lstlisting}

\subsubsection*{Initial Operator/Problem-Space Characterization}
\begin{lstlisting}[breaklines]
The problem space for delivering a specific amount of flucotone using flucotone-holding devices (FHDs) can be characterized by the following operators, preconditions, and postconditions:

### Operators

1. **Fill (FHDx)**: Completely fills FHDx from the power plant.
   - **Preconditions**: FHDx is not full.
   - **Postconditions**: FHDx is full (FHDx = capacity of FHDx).

2. **Empty (FHDx)**: Empties FHDx.
   - **Preconditions**: FHDx is not empty.
   - **Postconditions**: FHDx is empty (FHDx = 0).

3. **Transfer (FHDx, FHDy)**: Transfers flucotone from FHDx to FHDy until FHDx is empty or FHDy is full.
   - **Preconditions**: FHDx is not empty, and FHDy is not full.
   - **Postconditions**: FHDx decreases by the transferred amount, and FHDy increases by the same amount, without exceeding FHDy's capacity.

### Formal Notation

- Let \(FHDx\) and \(FHDy\) represent the state of two flucotone-holding devices, where \(x\) and \(y\) are their respective capacities.
- \(Fill(FHDx) \rightarrow (FHDx = x)\) when \(FHDx < x\).
- \(Empty(FHDx) \rightarrow (FHDx = 0)\) when \(FHDx > 0\).
- \(Transfer(FHDx, FHDy) \rightarrow (FHDx' = max(0, FHDx - (y - FHDy)), FHDy' = min(y, FHDy + FHDx))\) when \(FHDx > 0\) and \(FHDy < y\).

These operators, along with their preconditions and postconditions, define the actions that can be taken within the problem space to manipulate the state of the FHDs towards achieving the goal of delivering a specific amount of flucotone.
\end{lstlisting}

\subsubsection*{Refinement of Operator/Problem-Space Characterization}
\begin{lstlisting}[breaklines]
Upon reviewing the initial problem space characterization, the following improvements are suggested to ensure completeness and accuracy:

### Revised Operators with Preconditions and Postconditions

1. **Fill(FHDx)**
   - **Preconditions**: \(FHDx < Capacity_{FHDx}\)
   - **Postconditions**: \(FHDx = Capacity_{FHDx}\)

2. **Empty(FHDx)**
   - **Preconditions**: \(FHDx > 0\)
   - **Postconditions**: \(FHDx = 0\)

3. **Transfer(FHDx, FHDy)**
   - **Preconditions**: \(FHDx > 0\) AND \(FHDy < Capacity_{FHDy}\)
   - **Postconditions**: \(FHDx' = max(0, FHDx - (Capacity_{FHDy} - FHDy))\), \(FHDy' = min(Capacity_{FHDy}, FHDy + FHDx)\)

### Corrections and Additions

- **Clarification on Transfer Operator**: The original characterization did not explicitly account for the exact amount transferred. The revised postconditions now clearly define the new states of \(FHDx\) and \(FHDy\) after the transfer, ensuring that the transfer does not exceed \(FHDy\)'s capacity or result in \(FHDx\) having a negative amount of flucotone.
- **Addition of Preconditions**: Explicitly stating that \(FHDx\) must not be empty for a transfer to occur and \(FHDy\) must not be full ensures that the operators are only applied when their actions are possible within the physical constraints of the problem.
- **Specification of Capacities**: The use of \(Capacity_{FHDx}\) and \(Capacity_{FHDy}\) in preconditions and postconditions clarifies that each FHD has a specific capacity, which is crucial for understanding the limits within which the operators can function.

These revisions aim to provide a more precise and comprehensive characterization of the problem space, ensuring that the operators, along with their preconditions and postconditions, accurately reflect the actions that can be taken to manipulate the state of the FHDs towards achieving the goal of delivering a specific amount of flucotone.
\end{lstlisting}

\subsubsection*{Generation of Search Control}
\begin{lstlisting}[breaklines]
Undesirable states and sequences in the context of this problem include:

1. **Undesirable States:**
   - Any state where the total amount of flucotone in both FHDs does not contribute to achieving the exact goal amount (e.g., states where the sum of flucotone in FHD1 and FHD2 is greater than the goal or cannot be combined to reach the goal due to the capacities of the FHDs).
   - States where both FHDs are partially filled but their combined flucotone does not equal the goal and cannot be adjusted to meet the goal due to the lack of measurement tools.

2. **Undesirable Sequences of Operator Applications:**
   - Sequences that involve filling an FHD and then immediately emptying it without transferring any flucotone to the other FHD or towards achieving the goal (e.g., \(Fill(FHDx) \rightarrow Empty(FHDx)\)).
   - Sequences that result in a loop, where the same set of operations is repeated without making progress towards the goal (e.g., \(Fill(FHD1) \rightarrow Transfer(FHD1, FHD2) \rightarrow Empty(FHD2) \rightarrow Transfer(FHD2, FHD1)\), repeated without reaching the goal state).
   - Sequences where flucotone is transferred back and forth between FHDs without getting closer to the goal amount (e.g., \(Transfer(FHD1, FHD2) \rightarrow Transfer(FHD2, FHD1)\) without any net gain towards achieving the goal).

Formally, undesirable states can be characterized as any state \(S\) where \(FHD1 + FHD2 \neq Goal\) and no sequence of valid operations can transform \(S\) into a goal state \(SG\) where \(FHD1 + FHD2 = Goal\). Undesirable sequences can be characterized by any series of operations \(O_1, O_2, ..., O_n\) where the application of \(O_n\) directly negates the effect of \(O_{n-1}\) or fails to make progress towards \(SG\).
\end{lstlisting}

\end{document}